\documentclass[10pt,twocolumn,letterpaper]{article}

\usepackage[pagenumbers]{cvpr} 

\usepackage{algorithm}
\usepackage{algpseudocode}   
\algrenewcommand{\algorithmiccomment}[1]{\hfill{\color{gray}$\triangleright$\,#1}} 

\usepackage{booktabs}
\usepackage{multirow}
\usepackage{graphicx}

\definecolor{cvprblue}{rgb}{0.21,0.49,0.74}
\usepackage[pagebackref,breaklinks,colorlinks,allcolors=cvprblue]{hyperref}

\def\confName{CVPR}
\def\confYear{2027}

\title{AngularFM: Flow Matching on Joint Circles for Robot Manipulation}

\author{
Jiahang Tu$^{1,2}$\thanks{This work was conducted during the author's internship at KlingAI.}
\quad Yucheng Wang$^{2,3}$
\quad Yuzhuo Chen$^{2}$
\quad Xiaohan Li$^{2}$
\quad Xinyu Wang$^{2,4}$ \\[0.5em]
Yue Ma$^{3}$
\quad Liwen Mei$^{1}$
\quad Wenhu Zhang$^{3}$
\quad Fengyu Yang$^{5}$ 
\quad Sifan Zhou$^{6}$ \\[0.5em]
Guoxin Zhang$^{2}$
\quad Pengfei Wan$^{2}$
\quad Hanbin Zhao$^{1}$\thanks{Corresponding author.}
\quad Hui Qian$^{1}$\\[0.8em]
$^{1}$Zhejiang University
\quad $^{2}$KlingAI Research\\
$^{3}$The Hong Kong University of Science and Technology\\
$^{4}$Tsinghua University
\quad $^{5}$Yale University
\quad $^{6}$Southeast University\\[0.3em]
{\small \textbf{Project Page:}
\url{https://ttuhamg.github.io/AngularFM/}}
}

\begin{document}
\maketitle
\begin{abstract}
Many robot policies, including vision-language-action models and world action models, generate actions using diffusion or flow matching.
For a fixed robot, generating joint-position targets avoids inverse kinematics.
Revolute-joint angles can be represented on the unit circle $\mathbb{S}^1$, motivating three design criteria: continuous representations, consistency with circular geometry, and manifold-preserving generation.
The raw scalar representation cuts the circle open and maps it to an interval on the real line, introducing a discontinuity and distorting angular distances.
The sine–cosine representation restores continuity, but unconstrained Euclidean flow matching can produce off-manifold predictions that decode to incorrect joint angles.
We therefore propose AngularFM, a flow matching method that represents each joint angle as a phase on $\mathbb{S}^1$ and learns an angular velocity field over flow time to generate joint-position targets. 
Starting from noise sampled around the current joint angles, AngularFM evolves each phase along its circle, satisfying all three design criteria by construction.
We introduce a suite of rotation-sensitive manipulation tasks and show that AngularFM improves the overall success rate. 
It also achieves high success rates on LIBERO and RoboTwin 2.0 and outperforms baselines on six real-world tasks using Franka and AgileX robots.
\end{abstract}    
\section{Introduction}
\label{sec:intro}

Diffusion and flow matching are increasingly used to generate robot actions, including in diffusion policies \cite{chi2025diffusion}, vision-language-action models \cite{intelligence2025pi_, brohan2023rt}, and world action models \cite{ye2026world, zhu2025unified}. A common design choice for policies that predict position targets is whether to output end-effector poses or joint positions. For a fixed robot, directly generating joint-position targets avoids the inverse-kinematics step involved in converting end-effector pose targets into joint commands. A recent empirical study \cite{feng2026demystifying} also reports particular benefits of joint-space actions for flow matching policies. In this paper, we focus on how the representation and generation of these joint targets should reflect their underlying geometry.

For revolute joints, this geometry reflects the periodic structure of rotation. In joints that permit multiple turns, such as the UR3e’s \cite{universalrobots2024ur3e} final wrist joint and those in RoboTwin 2.0 \cite{chen2025robotwin} simulation setting, angles separated by $2\pi$ share the same rotational phase. These phases can be naturally represented on the unit circle $\mathbb{S}^1$. We argue that this circular structure should guide how joint angles are encoded and how they are generated. An encoding should vary continuously with rotation, assign the same code to equivalent angles, and keep nearby phases close. In flow matching generation, samples should also remain on the circle as they evolve from noise into action targets. This motivates three design criteria for joint-angle generation: \textbf{continuous representations}, \textbf{consistency with circular geometry}, and \textbf{manifold-preserving generation}.

\begin{figure*}[t]
  \centering
   \includegraphics[width=1.0\linewidth]{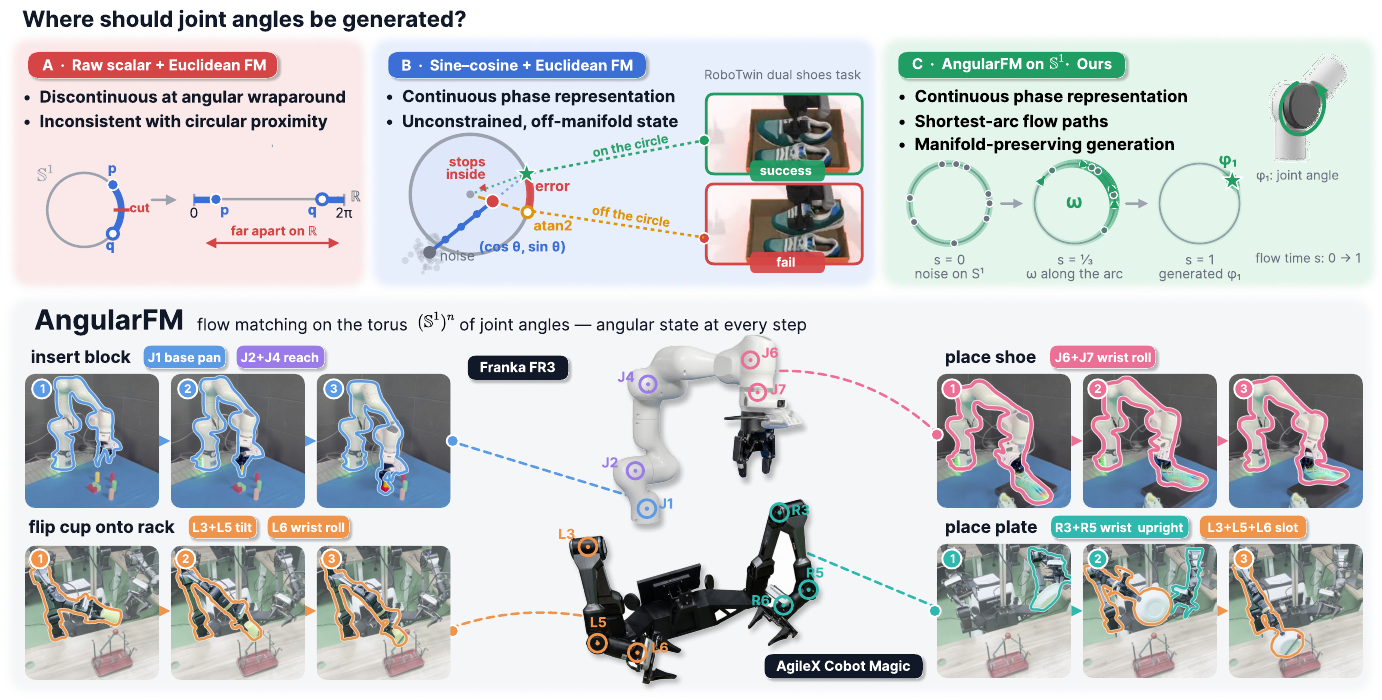}
    \vspace{-2em} 
   \caption{Joint-angle representations and generation on circular manifolds. Top: (A) Raw scalar representations can introduce discontinuities at angular wraparound, mapping nearby rotational phases to distant numerical values. (B) Sine--cosine encoding provides a continuous representation on the unit circle, but unconstrained Euclidean flow matching may produce off-manifold states that lead to angular decoding errors. (C) AngularFM represents joint angles as phases on $\mathbb{S}^1$ and learns an angular velocity field, preserving the circular manifold throughout generation. Bottom: Representative rotation-sensitive manipulation tasks on Franka FR3 and AgileX Cobot Magic, highlighting the joints involved in each task and their corresponding manipulation sequences.}
   \label{fig:teaser}
    \vspace{-5mm} 
\end{figure*}

These criteria reveal different limitations of two commonly used joint-angle representations, raw scalar and sine–cosine, when paired with Euclidean flow matching. The raw scalar representation models joint angles on the real line, where equivalent rotational phases can receive labels separated by full turns. Wrapping these angles into a fixed interval assigns each phase a single scalar representative, but introduces a discontinuity at the wraparound boundary \cite{zhou2019continuity}, as illustrated in Fig.~\ref{fig:teaser}A. Phases across this boundary can be close on the circle yet far apart numerically, so their Euclidean distance in the scalar representation does not reflect their circular proximity. Thus, the raw scalar representation can violate the first two criteria. In contrast, the sine–cosine representation provides a continuous embedding of rotational phases on the unit circle~\cite{beyer2015biternion,zhou2019continuity}. However, when paired with unconstrained Euclidean flow matching, sine--cosine states may deviate from the unit circle during generation, potentially introducing angular decoding errors, as illustrated in Fig.~\ref{fig:teaser}B. Therefore, while the sine–cosine representation satisfies the first two criteria, its Euclidean generative process does not guarantee manifold-preserving generation.

To satisfy these three criteria, we propose AngularFM, a flow matching method for joint-position generation on joint circles, as illustrated in Fig.~\ref{fig:teaser}C. AngularFM retains the continuous sine–cosine representation of each joint phase but replaces unconstrained planar velocities with a scalar angular velocity for each joint and future action step. We derive this parameterization by enforcing the unit-circle constraint, which eliminates radial motion and leaves only tangential evolution. The angular velocity field is trained using displacements along shortest circular arcs and integrated through rotations, preserving the manifold at every sampling step. To incorporate the robot’s current proprioception, we further initialize future joint phases through angular perturbations around the corresponding current joint angles. This state-anchored source provides a shared reference for the action chunk while retaining stochastic generation. Together, continuous phase inputs, shortest-arc training paths, and rotation-based sampling realize continuous representations, consistency with circular geometry, and manifold-preserving generation, respectively.

We evaluate AngularFM against raw scalar and sine–cosine representations using the same policy backbone, training data, and training budget. Since aggregate results on general-purpose manipulation benchmarks may obscure the challenges posed by large joint rotations and angular wraparound, we introduce a targeted suite of seven rotation-sensitive manipulation tasks in RoboTwin 2.0 \cite{chen2025robotwin}, comprising 2,800 demonstrations. Across multiple training seeds and scene sets, AngularFM achieves higher success rates than both baselines. Ablations further demonstrate the benefit of anchoring the source distribution at the current joint configuration. Beyond this controlled suite, AngularFM achieves strong performance on LIBERO \cite{liu2023libero} and the full fifty-task RoboTwin 2.0 benchmark. It also outperforms the baselines on six real-world tasks using a single-arm Franka and a dual-arm AgileX robot. These results support aligning both the representation and generation of joint angles with circular geometry.

Our contributions are threefold.
\begin{itemize}
    \item We formulate three design criteria for joint-angle generation and identify complementary limitations of existing representations.

    \item We propose AngularFM, a manifold-preserving flow matching method on joint circles with an angular velocity field and a state-anchored source distribution.

    \item We introduce a rotation-sensitive task suite and validate AngularFM through controlled comparisons, standard benchmarks, and real-world robot experiments.
\end{itemize}
\section{Related Work}
\label{sec:related_work}

\paragraph{Vision-Language-Action and World Action Models.}
Vision-language-action models integrate visual observations and language instructions with action prediction, often leveraging pretrained vision-language models and heterogeneous robot data \cite{brohan2023rt,kim2024openvla,black2024pi_0,intelligence2025pi_,li2024cogact,bjorck2025gr00t}.
For example, $\pi_0$ \cite{black2024pi_0} combines a pretrained vision-language model with flow matching to generate continuous action chunks, while GR00T N1 \cite{bjorck2025gr00t} couples a vision-language module with a diffusion transformer for humanoid control.
World action models leverage predictive visual dynamics for policy learning through joint video--action modeling or predictive visual representations~\cite{zhu2025unified,hu2024video,ye2026world,yuan2026fast}.
UWM~\cite{zhu2025unified} integrates video and action diffusion with modality-specific diffusion timesteps, supporting policy and dynamics learning within a shared model.
Fast-WAM~\cite{yuan2026fast} separates video co-training from test-time future generation and enables action-only inference while retaining the benefits of predictive training.
These works advance multimodal policy learning through pretrained representations and predictive objectives.
AngularFM addresses a complementary problem by aligning the representation and generation of joint angles with circular geometry within such policy frameworks.

\paragraph{Action Spaces for Robot Manipulation.}
Robot policies commonly predict either end-effector motions or joint-position targets.
End-effector action spaces range from restricted 4-DoF motions, such as 3D translation and yaw rotation in QT-Opt \cite{jang2018qt}, to full 6-DoF poses, as in PerAct \cite{shridhar2023perceiver} and 3D Diffuser Actor \cite{ke20243d}.
The orientation component of a full 6-DoF pose can be encoded using unit quaternions or continuous 6D rotation representations \cite{zhou2019continuity}.
Alternatively, policies can directly predict joint-position targets, avoiding the inverse-kinematics conversion from end-effector poses.
Joint-level control is particularly relevant to highly articulated or redundant systems, such as dexterous hands and mobile manipulators, where a single end-effector pose cannot fully specify the robot configuration \cite{handa2020dexpilot,fu2024mobile}.
A large-scale empirical study \cite{feng2026demystifying} further shows that joint-space actions favor control stability for fixed robot embodiments, particularly with flow matching policies, while end-effector-space actions offer advantages in cross-embodiment generalization.
Within joint-position prediction, we study how to represent and generate the one-dimensional rotational phase of each revolute joint while jointly modeling the full multi-joint action chunk.

\paragraph{Rotation Representations and Manifold-Based Generation.}
Biternion networks \cite{beyer2015biternion} provide a continuous sine--cosine encoding for periodic angles.
Such an encoding specifies how an angle is represented, but does not by itself constrain the dynamics used to generate it.
Riemannian Flow Matching~\cite{chen2024flow} learns continuous normalizing flows \cite{papamakarios2021normalizing} on manifolds, including tori and product spaces.
RFMP \cite{braun2024riemannian} applies this framework to observation-conditioned robot motion learning, while subsequent visuomotor policies \cite{ding2025fast} study manifold-valued manipulation actions, including end-effector orientations represented by unit quaternions. In contrast to these end-effector formulations, AngularFM directly generates joint-position action chunks through
angular velocity fields on joint circles. Our controlled comparisons distinguish the effects of continuous representations and manifold-preserving generation in joint-space control.

\section{Method}
\label{sec:method}

AngularFM aligns joint-angle representations and generative dynamics with circular geometry.
We first examine Euclidean flow matching with the sine--cosine representation.
We then derive angular velocity prediction and rotation-based sampling on joint circles.
Algorithms~\ref{alg:afm_train_final} and~\ref{alg:afm_sample_final} summarize the training and inference procedures.

\paragraph{Problem Formulation.}
Let $o=(I,\ell,q_{\mathrm{cur}})$ contain images $I$, a language instruction $\ell$, and current joint positions $q_{\mathrm{cur}}\in\mathbb R^n$.
We learn from demonstrations $(o,Q_1)\sim\mathcal D$, where $Q_1\in\mathbb R^{H\times n}$ contains $H$ future targets for $n$ revolute joints.
We write $q_s^{h,i}$ for an angular coordinate of joint $i$ at future action step $h$ and flow time $s\in[0,1]$.
The subscripts $0$ and $1$ denote the noise and data ends of flow time, not execution steps.
For a fixed future step, we omit $h$ and write $q_s^i$.

The sine--cosine map $\mathbf u(q)=(\cos q,\sin q)^\top$ gives the phase representation $\mathbf z_s^{h,i}=\mathbf u(q_s^{h,i})\in\mathbb S^1$.
Equivalent angular representatives have the same phase.
Stacking these phases gives $\mathbf Z_s\in\mathcal M$, where $\mathcal M=(\mathbb S^1)^{H\times n}$ is represented by an $H\times n\times2$ array.
The network jointly models all joints and future steps.
Gripper commands and other linear components retain Euclidean flow matching.

\subsection{Euclidean Flow Matching with Sine--Cosine Encoding}
\label{sec:afm_euclidean}

For a fixed future step, the demonstrated joint target $q_1^i$ is encoded as $\mathbf z_1^i=\mathbf u(q_1^i)$.
Euclidean flow matching~\cite{lipman2022flow} samples $\mathbf x_0^i\sim\mathcal N(0,I_2)$, with $I_2$ the two-dimensional identity matrix, and constructs
\begin{equation}
\mathbf x_s^i=(1-s)\mathbf x_0^i+s\mathbf z_1^i,
\qquad
\dot{\mathbf x}_s^i=\mathbf z_1^i-\mathbf x_0^i.
\label{eq:afm_efm_path}
\end{equation}
The dot denotes differentiation with respect to flow time.
Unlike $\mathbf z_s^i$, the planar state $\mathbf x_s^i$ is not restricted to the unit circle.

For the full chunk, the network predicts a planar velocity field $\mathbf V_\theta(\mathbf X_s,s,o)\in\mathbb R^{H\times n\times2}$ with parameters $\theta$.
Here, $\mathbf X_s$ stacks the planar states across future steps and joints.
The training objective is
\begin{equation}
\mathcal L_{\mathrm E}(\theta)
=\mathbb E\!\left[
 w(s)\left\|\mathbf V_\theta(\mathbf X_s,s,o)
 -(\mathbf Z_1-\mathbf X_0)\right\|_F^2\right].
\label{eq:afm_efm_loss}
\end{equation}
The expectation is over demonstrations, Gaussian source noise, and flow times $s\sim p$, where $p$ is the training distribution over $[0,1]$.
The prescribed weight $w(s)>0$ scales the loss at each flow time.

Inference integrates the learned field to obtain $\widehat{\mathbf X}_1$.
Although the clean targets $\mathbf Z_1$ satisfy the unit-circle constraint, neither the Gaussian source nor the linear training paths are confined to the circle.
The generated endpoint $\widehat{\mathbf X}_1$ can also remain off-manifold after finite-step sampling. As shown in Fig.~\ref{fig:teaser}B, recovering joint angles through $\operatorname{atan2}$ can amplify prediction errors, leading to inaccurate angular estimates. We analyze this decoding sensitivity and the associated off-manifold deviations in
Appendix~\ref{sec:supp_offcircle}. Thus, continuous sine--cosine representations do not guarantee manifold-preserving generation.

\subsection{Angular Flow Matching on Joint Circles}
\label{sec:afm_velocity}

AngularFM places the source and training paths on joint circles and restricts the learned velocity field to their tangent directions.
The sine--cosine input representation is retained.

\paragraph{Angular Velocity Field.}
Let $K=\left[\begin{smallmatrix}0&-1\\1&0\end{smallmatrix}\right]$ rotate a planar vector counterclockwise by $\pi/2$.
At each phase $\mathbf z_s^i=\mathbf u(q_s^i)$, the vectors $\mathbf z_s^i$ and $K\mathbf z_s^i$ form an orthonormal basis of the ambient plane, corresponding to the radial and tangential directions. Any planar velocity can therefore be decomposed into $\mathbf v_s^i = \alpha_s^i\mathbf z_s^i+\omega_s^iK\mathbf z_s^i$, where $\alpha_s^i$ and $\omega_s^i$ are its radial and tangential coefficients.

Enforcing the unit-circle constraint removes the radial component $\alpha_s^i=0$, leaving only tangential motion $\mathbf v_s^i = \omega_s^iK\mathbf z_s^i$.
Differentiating the sine--cosine map then gives
\begin{equation}
\begin{aligned}
\dot{\mathbf z}_s^i
& = \frac{\partial\mathbf u(q_s^i)}{\partial q_s^i}
  \frac{dq_s^i}{ds} = \dot q_s^i K\mathbf z_s^i \\
& =\omega_s^i K\mathbf z_s^i,
\quad \Rightarrow \quad 
\dot q_s^i=\omega_s^i.
\label{eq:afm_angular_ode}
\end{aligned}
\end{equation}
Thus, $\omega_s^i$ is the angular velocity over flow time.

For the full action chunk, AngularFM predicts $\boldsymbol\omega_\theta(\mathbf Z_s,s,o) \in\mathbb R^{H\times n}$.
Each entry $[\boldsymbol\omega_\theta(\mathbf Z_s,s,o)]_{h,i}$ specifies the angular velocity of joint $i$ at future action step $h$. Compared with the Euclidean velocity field $\mathbf V_\theta$, AngularFM predicts one intrinsic angular velocity instead of two unconstrained planar velocity components while retaining continuous phase inputs.

\paragraph{State-Anchored Source Distribution.}
Flow matching requires a source distribution from which generation begins. While a uniform distribution on \(\mathbb{S}^1\) satisfies the unit-circle constraint, it does not incorporate the robot's current joint configuration. Since future joint-position targets are conditioned on the current robot state, we initialize the flow around the current joint angles. Specifically, we sample
\begin{equation}
q_0^{h,i}=q_{\mathrm{cur}}^i+\epsilon^{h,i},
\qquad
\epsilon^{h,i}\overset{\mathrm{i.i.d.}}{\sim}
\mathcal N(0,\kappa^{-1}).
\label{eq:afm_source}
\end{equation}

The corresponding phases \(\mathbf z_0^{h,i}=\mathbf u(q_0^{h,i})\) define a state-anchored wrapped normal source distribution on the joint circles. Here, \(\kappa>0\) controls the angular-noise precision, with a perturbation standard deviation of \(\kappa^{-1/2}\) radians. All future steps of a joint share its current phase as the source center, while retaining independent stochastic perturbations.

\begin{algorithm}[t]
\caption{AngularFM training}
\label{alg:afm_train_final}
\small
\begin{algorithmic}[1]
\raggedright
\Statex \textbf{Input:} Demonstrations $(o,Q_1)\sim\mathcal D$,
\Statex \quad $o=(I,\ell,q_{\mathrm{cur}})$,
  $Q_1\in\mathbb R^{H\times n}$;
\Statex \quad network parameters $\theta$,
  noise precision $\kappa>0$,
\Statex \quad flow-time distribution $p$,
  loss weight $w$.
\Statex \textbf{Output:}
  Trained angular velocity field $\boldsymbol\omega_\theta$.
\Statex \textbf{Operators:}
  $\mathbf u(a)=(\cos a,\sin a)^\top$;
\Statex \quad $\operatorname{ang}(\mathbf u(a),\mathbf u(b))$
\Statex \qquad
  $=\operatorname{atan2}(\sin(b-a),\cos(b-a))$.
\Statex \quad Both operators act elementwise on joint arrays.
\Statex \quad For antipodal phases,
  $\operatorname{ang}$ returns $-\pi$.
\Repeat
  \State Sample demonstration $(o,Q_1)\sim\mathcal D$.
  \State Sample
    $\epsilon^{h,i}\overset{\mathrm{i.i.d.}}{\sim}
    \mathcal N(0,\kappa^{-1})$
    \Comment{$h=1{:}H$, $i=1{:}n$}
  \State $Q_0\gets
    [q_{\mathrm{cur}}^i+\epsilon^{h,i}]_{h,i}$
    \Comment{source angles, $H\times n$}
  \State $\boldsymbol\Delta\gets
    \operatorname{ang}(\mathbf u(Q_0),\mathbf u(Q_1))$
    \Comment{shortest angular displacements}
  \State Sample flow time $s\sim p$.
  \State $\mathbf Z_s\gets
    \mathbf u(Q_0+s\boldsymbol\Delta)$
    \Comment{circle-valued conditional path}
  \State $\widehat{\boldsymbol\omega}\gets
    \boldsymbol\omega_\theta(\mathbf Z_s,s,o)$
  \State $\mathcal L\gets w(s)
    \|\widehat{\boldsymbol\omega}-\boldsymbol\Delta\|_F^2$
  \State Update $\theta$ using $\nabla_\theta\mathcal L$.
\Until{training is complete}
\end{algorithmic}
\end{algorithm}

\begin{algorithm}[t]
\caption{AngularFM sampling and joint-target reconstruction}
\label{alg:afm_sample_final}
\small
\begin{algorithmic}[1]
\raggedright
\Statex \textbf{Input:}
  Observation $o=(I,\ell,q_{\mathrm{cur}})$,
\Statex \quad trained angular velocity field
  $\boldsymbol\omega_\theta$,
\Statex \quad noise precision $\kappa>0$,
  $H$ future steps,
\Statex \quad $n$ joints,
  flow-time grid $0=s_0<\cdots<s_N=1$.
\Statex \textbf{Output:}
  Joint-position chunk $\bar Q\in\mathbb R^{H\times n}$.
\State Sample
  $\epsilon^{h,i}\overset{\mathrm{i.i.d.}}{\sim}
  \mathcal N(0,\kappa^{-1})$
  \Comment{$h=1{:}H$, $i=1{:}n$}
\State $\widehat Q_{s_0}\gets
  [q_{\mathrm{cur}}^i+\epsilon^{h,i}]_{h,i}$
  \Comment{state-anchored initial angles}
\For{$k=0,\ldots,N-1$}
  \State $\mathbf Z_k\gets\mathbf u(\widehat Q_{s_k})$
    \Comment{phases on $\mathbb S^1$}
  \State $\boldsymbol\omega_k\gets
    \boldsymbol\omega_\theta(\mathbf Z_k,s_k,o)$
  \State $\Delta s_k\gets s_{k+1}-s_k$
  \State $\widehat Q_{s_{k+1}}\gets
    \widehat Q_{s_k}+\Delta s_k\boldsymbol\omega_k$
    \Comment{on-circle rotation}
\EndFor
\State $\widehat Q_1\gets\widehat Q_{s_N}$
  \Comment{generated phase angles}
\State $\bar q^{0,:}\gets q_{\mathrm{cur}}$
  \Comment{anchor the execution branch}
\For{$h=1,\ldots,H$}
  \State $\mathbf d^h\gets\operatorname{ang}\!\left(
    \mathbf u(\bar q^{h-1,:}),
    \mathbf u(\hat q_1^{h,:})\right)$
    \Comment{nearest angular increment}
  \State $\bar q^{h,:}\gets\bar q^{h-1,:}+\mathbf d^h$
\EndFor
\State \Return $\bar Q=[\bar q^{h,i}]_{h,i}$
\end{algorithmic}
\end{algorithm}

\paragraph{Shortest-Arc Training Paths.}
Given the source and target phases, flow matching requires
a conditional path connecting them.
As shown in Sec.~\ref{sec:afm_euclidean}, Euclidean linear
interpolation can leave the unit circle even when both
endpoints lie on it.
To ensure consistency with circular geometry, we instead
construct conditional paths along the shortest circular arcs.

For unit phases $\mathbf a,\mathbf b\in\mathbb S^1$,
let $\operatorname{ang}(\mathbf a,\mathbf b)$ denote their
signed shortest angular displacement, computed as
$\operatorname{atan2}((K\mathbf a)^\top\mathbf b,
\mathbf a^\top\mathbf b)$.
For each source and target phase, we define
\begin{equation}
\begin{aligned}
\Delta^{h,i}
&=\operatorname{ang}\!\left(
\mathbf u(q_0^{h,i}),\mathbf u(q_1^{h,i})\right)\,,\\
\mathbf z_s^{h,i}
&=\mathbf u\!\left(q_0^{h,i}+s\Delta^{h,i}\right).
\end{aligned}
\label{eq:afm_path}
\end{equation}
These paths follow the shortest geodesics on $\mathbb S^1$,
preserving the unit-circle constraint throughout flow time.
Differentiating the path gives
$\dot{\mathbf z}_s^{h,i}
=\Delta^{h,i}K\mathbf z_s^{h,i}$.
Thus, $\Delta^{h,i}$ provides the constant conditional
angular velocity along each training path.

\paragraph{Angular Flow Matching Objective.}
Conditional flow matching learns a velocity field by
matching the velocities of sampled conditional
paths~\cite{lipman2022flow,chen2024flow}.
Let $\boldsymbol\Delta\in\mathbb R^{H\times n}$
collect the target angular velocities.
Both the predicted and target velocities lie along
the unit tangent $K\mathbf z_s^{h,i}$.
Since $\|K\mathbf z_s^{h,i}\|_2=1$,
the tangent-velocity matching error reduces to
the squared difference between their angular velocities.
The resulting objective is
\begin{equation}
\mathcal L_{\mathrm{AngularFM}}(\theta)
=\mathbb E\!\left[
w(s)\left\|
\boldsymbol\omega_\theta(\mathbf Z_s,s,o)
-\boldsymbol\Delta
\right\|_F^2\right].
\label{eq:afm_loss}
\end{equation}
Thus, AngularFM learns intrinsic angular velocities
without introducing radial velocity components.
Algorithm~\ref{alg:afm_train_final} summarizes
the complete training procedure.

\subsection{Rotation-Based Sampling and Joint-Position Targets}
\label{sec:afm_sampling}

\paragraph{Rotation-Based Sampling.}
Although the learned angular velocity field is tangent to the joint circles, discrete integration must also preserve the unit-circle constraint. Directly applying an Euler step to the planar phase coordinates can move the state off the circle. We therefore integrate the predicted angular velocities through rotations, ensuring manifold-preserving generation at every sampling step.

Let $0=s_0<\cdots<s_N=1$ denote $N$ sampling steps, with $\Delta s_k=s_{k+1}-s_k$. We initialize the sampled angles as $\widehat Q_{s_0}=Q_0$ using the state-anchored source in Eq.~\eqref{eq:afm_source}. At each step $k$, the network predicts $\boldsymbol\omega_k= \boldsymbol\omega_\theta( \mathbf u(\widehat Q_{s_k}),s_k,o)$, where $\omega_k^{h,i}$ denotes the angular velocity of joint $i$ at future action step $h$. The sampling update is
\begin{equation}
\begin{aligned}
\hat q_{s_{k+1}}^{h,i}
&=\hat q_{s_k}^{h,i}
+\Delta s_k\omega_k^{h,i},\\
\mathbf u(\hat q_{s_{k+1}}^{h,i})
&=R(\Delta s_k\omega_k^{h,i})
\mathbf u(\hat q_{s_k}^{h,i}),
\end{aligned}
\label{eq:afm_sampling_update}
\end{equation}
where $R(\beta)=e^{\beta K}$ denotes a planar rotation by angle $\beta$. Since $R(\beta)$ is orthogonal, each update preserves the unit norm of the phase representation. Consequently, the source states, intermediate samples, and generated endpoints all remain on the joint circles without additional projection.

\paragraph{Joint-Position Reconstruction.}
While rotation-based sampling generates valid joint phases, the sine--cosine representation identifies angles that differ by integer multiples of $2\pi$. However, robot controllers require real-valued joint-position targets that maintain continuity along the action sequence. We therefore reconstruct the joint positions through sequential phase lifting.

Starting from the current joint position $\bar q^{0,i}=q_{\mathrm{cur}}^i$,
we recover each subsequent target as
\begin{equation}
    \bar q^{h,i}
=\bar q^{h-1,i}
+\operatorname{ang}\!\left(
\mathbf u(\bar q^{h-1,i}),
\mathbf u(\hat q_1^{h,i})
\right).
\end{equation}
This selects the angular representative nearest to the preceding joint-position target, preserving continuity across future action steps. The reconstructed targets form the action chunk $\bar Q\in\mathbb R^{H\times n}$. Algorithm~\ref{alg:afm_sample_final} summarizes the sampling and reconstruction procedure.

Together, continuous phase representations, shortest-arc conditional paths, and rotation-based sampling satisfy the three design criteria: \textbf{continuous representations}, \textbf{consistency with circular geometry}, and \textbf{manifold-preserving generation}.

\section{Experiments}
\label{sec:experiments}

Our experiments investigate whether aligning joint-angle representations and generative dynamics with circular geometry improves closed-loop robotic manipulation.
We first evaluate AngularFM on LIBERO \cite{liu2023libero} and RoboTwin 2.0 \cite{chen2025robotwin} to assess its performance across diverse manipulation tasks in Sec.~\ref{sec:simulation_benchmarks}). We then examine rotation-sensitive manipulation through task-subset analysis and controlled multi-seed comparisons in Sec.~\ref{sec:rotation_sensitive}. Finally, we validate AngularFM on physical robots in Sec.~\ref{sec:real_world} and conduct ablations to
assess the contribution of our modules in Sec.~\ref{sec:ablations}.

\subsection{Experimental Setup}
\label{sec:exp_setup}

\begin{table}[t]
\centering
\small
\caption{Reported success rates (\%) on LIBERO. Bold and underlined
averages indicate the highest and next-highest reported scores.}
\label{tab:libero}
\resizebox{1.0\columnwidth}{!}{%
\begin{tabular}{l|cccc|c}
\toprule
Method & Spatial & Object & Goal & Long & Average \\
\midrule
OpenVLA \cite{kim2024openvla} & 84.7 & 88.4 & 79.2 & 53.7 & 76.5 \\
UniVLA \cite{bu2505learning} & 96.5 & 96.8 & 95.6 & 92.0 & 95.2 \\
$\pi_0$ \cite{black2024pi_0} & 96.8 & 98.8 & 95.8 & 85.2 & 94.1 \\
$\pi_{0.5}$ \cite{intelligence2025pi_} & 98.8 & 98.2 & 98.0 & 92.4 & 96.9 \\
LingBot-VA \cite{li2026causal} & 98.5 & 99.6 & 97.2 & 98.5 & \textbf{98.5} \\
Motus \cite{bi2026motus} & 96.8 & 99.8 & 96.6 & 97.6 & 97.7 \\
Fast-WAM \cite{yuan2026fast} & 98.2 & 100.0 & 97.0 & 95.2 & 97.6 \\
Fast-WAM-Joint \cite{yuan2026fast} & 99.6 & 99.4 & 98.2 & 96.8 & \textbf{98.5} \\
CogVLA \cite{li2026cogvla} & 98.6 & 98.8 & 96.6 & 95.4 & 97.4 \\
DreamVLA \cite{zhang2026dreamvla} & 97.5 & 94.0 & 89.5 & 89.5 & 92.6 \\
CoT-VLA \cite{zhao2025cot} & 87.5 & 91.6 & 87.6 & 69.0 & 81.1 \\
\midrule
\textbf{AngularFM} & 99.0 & 99.4 & 97.0 & 97.2 & \underline{98.2} \\
\bottomrule
\end{tabular}}
\end{table}

\paragraph{LIBERO.}
We evaluate on LIBERO-Spatial, LIBERO-Object, LIBERO-Goal, and LIBERO-Long~\cite{liu2023libero}, covering forty tasks in total. We report success rates for each suite and their average.

\paragraph{RoboTwin 2.0.}
We evaluate on fifty bimanual manipulation tasks under clean and randomized conditions~\cite{chen2025robotwin}. The dataset contains 2,500 clean and 25,000 randomized demonstrations. We report average success rates for both conditions.

\paragraph{Rotation-sensitive manipulation tasks.}
Since aggregate benchmark scores can obscure the effect of large joint rotations and angular wraparound, we construct seven rotation-sensitive tasks in RoboTwin 2.0 based on wrist-joint angular variation and collect 2,800 new demonstrations in simulation.
Each method is trained with three random seeds and evaluated on three scene sets.
Task selection, data collection, and evaluation protocols are detailed in Appendix~\ref{sec:supp_suite}.

\paragraph{Real-world evaluation.}
We evaluate six rotation-sensitive tasks and provide examples in Fig.~\ref{fig:teaser}: shoe placement, block insertion, and block stacking on a single-arm Franka FR3, and cup inversion with handover, plate insertion, and bimanual rack flipping on a dual-arm AgileX Cobot Magic. They involve object reorientation, orientation alignment during placement, and coordinated bimanual rotation. We compare with $\pi_{0.5}$~\cite{intelligence2025pi_} and two Fast-WAM-Joint baselines using raw scalar and sine--cosine joint-angle representations, respectively.
Each method is evaluated over ten trials per task.

\paragraph{Implementation details.}
We adopt Fast-WAM-Joint~\cite{yuan2026fast} as our base architecture, because it outperformed Fast-WAM in our exploration. Our model uses the Wan2.2-5B backbone~\cite{wan2025wan} and a 1B-parameter action expert with a hidden dimension of 1024, totaling 6B parameters. Each chunk contains $H=32$ actions and nine video frames. We use $N=10$ sampling steps and a classifier-free guidance scale of 1.0. The angular-noise precision is $\kappa=2$, except in the main rotation-sensitive task comparison, where $\kappa=4$. For benchmark training, we use AdamW with a learning rate of $10^{-5}$, a global batch size of 256, and a weight decay of 0.01. The learning rate follows a cosine decay schedule with linear warm-up. Further training and implementation details are provided in Appendix~\ref{sec:supp_impl}.

\subsection{Evaluation on Simulation Benchmarks}
\label{sec:simulation_benchmarks}

\paragraph{LIBERO.}
As demonstrated in Tab.~\ref{tab:libero}, AngularFM achieves 99.0\%, 99.4\%, 97.0\%,
and 97.2\% across the four suites, yielding a 98.2\% average.
AngularFM performs comparably to the strongest entries in the table,
within 0.3 percentage points of the highest reported average.
These results demonstrate strong performance across general manipulation
tasks, although the aggregate scores provide limited separation among
the leading methods.

\paragraph{RoboTwin 2.0.}
On the full RoboTwin 2.0 benchmark in Tab.~\ref{tab:robotwin}, AngularFM achieves 95.20\% success in clean scenes and 94.24\% in randomized scenes, with a 94.7\% average. It ranks highest among the listed methods in both conditions
and outperforms other methods, like LingBot-VA by 2.5 percentage points on average. AngularFM also outperforms Fast-WAM-Joint by 4.1 percentage points in average success rate. These results demonstrate strong performance across diverse manipulation tasks. We next examine how joint-angle generation affects performance within RoboTwin 2.0.

\begin{table}[t]
\centering
\small
\caption{Reported success rates (\%) on RoboTwin 2.0. Bold and underlined averages indicate the highest and next-highest reported scores.}
\label{tab:robotwin}
\setlength{\tabcolsep}{4mm}{} 
\resizebox{1.0\columnwidth}{!}{%
\begin{tabular}{l|cc|c}
\toprule
Method & Clean & Randomized & Average \\
\midrule
X-VLA \cite{zheng2026x} & 72.80 & 72.84 & 72.8 \\
$\pi_0$ \cite{black2024pi_0} & 65.92 & 58.40 & 62.2 \\
$\pi_{0.5}$ \cite{intelligence2025pi_} & 82.74 & 76.76 & 79.8 \\
Motus \cite{bi2026motus} & 88.66 & 87.02 & 87.8 \\
LingBot-VA \cite{li2026causal} & 92.90 & 91.50 & 92.2 \\
Fast-WAM \cite{yuan2026fast} & 91.88 & 91.78 & 91.8 \\
Fast-WAM-Joint \cite{yuan2026fast} & 90.84 & 90.32 & 90.6 \\
ABot-M0 \cite{yang2026abot} & 86.06 & 85.08 & 85.6 \\
ABot-M0.5 \cite{chen2026abot} & 94.00 & 94.20 & \underline{94.1} \\
Qwen-VLA \cite{wang2026qwen} & 86.10 & 87.20 & 86.7 \\
Qwen-RobotManip \cite{yuan2026qwen} & 93.70 & 94.00 & 93.9 \\
ImageWAM \cite{zhang2026imagewam} & 93.20 & 93.56 & 93.4 \\
AHA-WAM \cite{cai2026aha} & 93.40 & 92.20 & 92.8 \\
Fast-VLA \cite{gong2026fast} & 92.46 & 92.40 & 92.4 \\
Flash-WAM \cite{akbari2026flash} & 88.42 & 82.66 & 85.5 \\
\midrule
\textbf{AngularFM} & 95.20 & 94.24 & \textbf{94.7} \\
\bottomrule
\end{tabular}}
\end{table}

\begin{figure}[t]
  \centering
   \includegraphics[width=1.0\columnwidth]{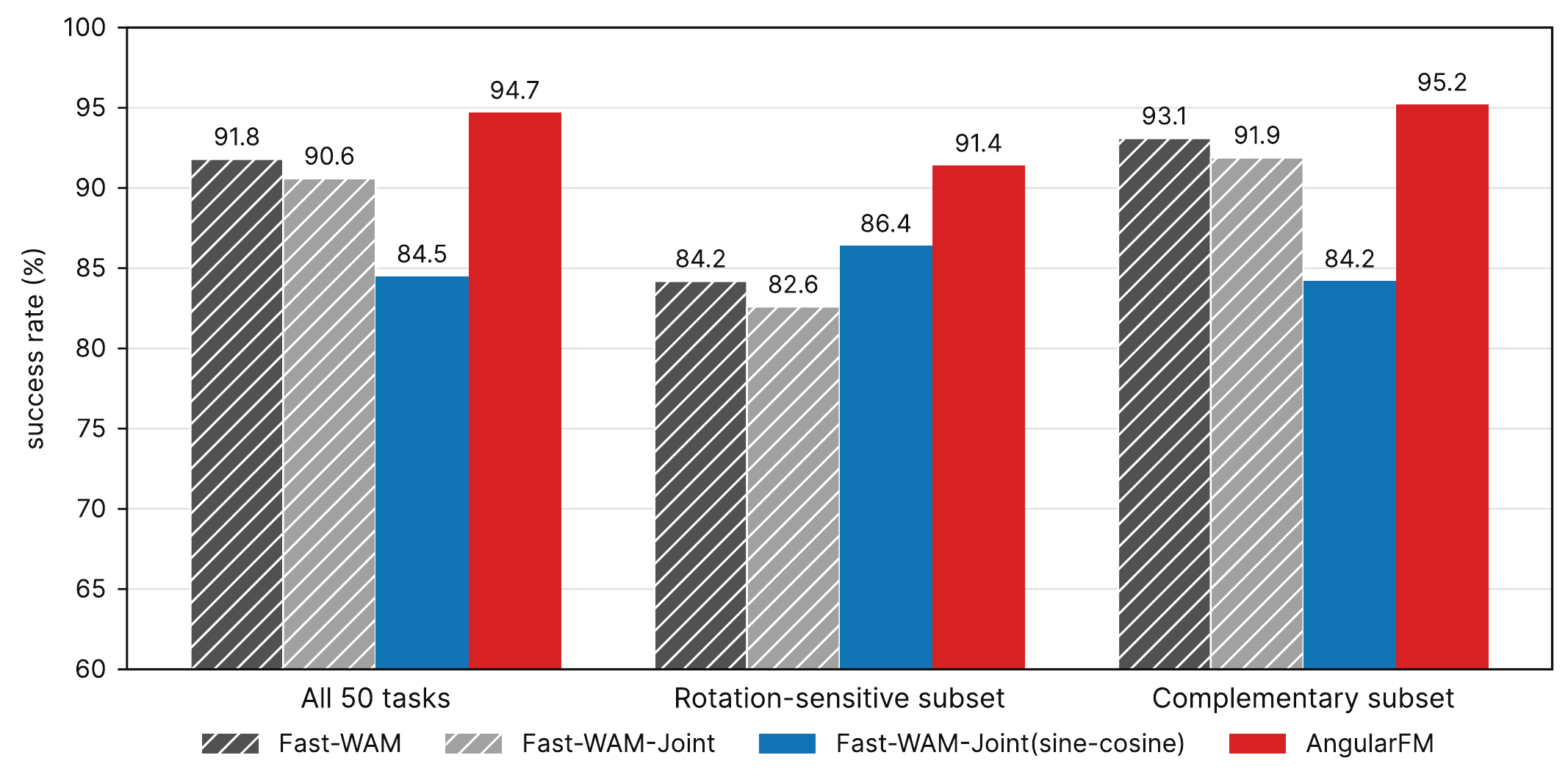}
    \vspace{-1em} 
    \caption{
    Success rates on RoboTwin 2.0 across all 50 tasks, the rotation-sensitive subset (7 tasks), and the complementary subset (43 tasks).
    }
    \label{fig:robotwin_groups}
    \vspace{-5mm} 
\end{figure}

\subsection{Rotation-Sensitive Manipulation Tasks}
\label{sec:rotation_sensitive}

\begin{figure*}[t]
  \centering
   \includegraphics[width=1.0\linewidth]{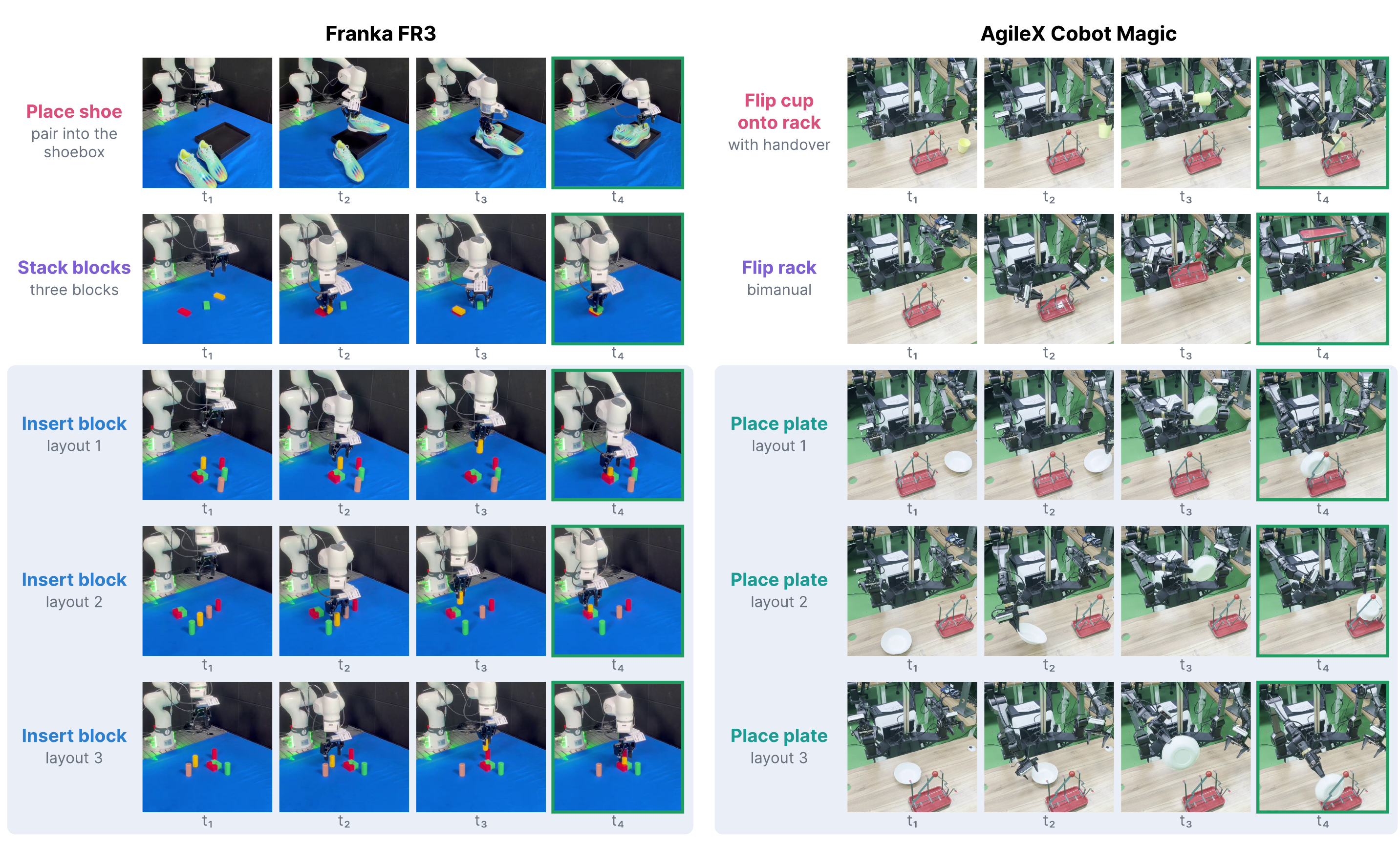}
    \vspace{-1em} 
   \caption{Execution process of AngularFM on real-world robotic manipulation tasks. Left: Franka FR3 performs shoe placement, block stacking, and block insertion under three object layouts. Right: AgileX Cobot Magic performs cup inversion with handover, bimanual rack flipping, and plate insertion under three object layouts. Each row shows four representative frames ($t_1$--$t_4$) from a successful manipulation sequence, with the final state highlighted in green.}
   \label{fig:real_deployment}
    \vspace{-5mm} 
\end{figure*}

\paragraph{Task-subset analysis.}
Fig.~\ref{fig:robotwin_groups} compares performance across the rotation-sensitive subset and its complementary subset of RoboTwin 2.0. We additionally train a Fast-WAM-Joint variant using sine--cosine action representations with Euclidean flow matching to compare against its original raw-scalar formulation (Fast-WAM-Joint). On the rotation-sensitive subset, sine--cosine improves success from 82.6\% to 86.4\%, also surpassing Fast-WAM (84.2\%). This suggests that continuous circular representations benefit rotation-sensitive tasks. However, on the complementary subset, sine--cosine achieves only 84.2\%, falling 7.7 percentage points below raw-scalar Fast-WAM-Joint (91.9\%).
This indicates that representation continuity alone is insufficient when paired with unconstrained Euclidean flow matching. AngularFM keeps the continuous representation, generates on the joint circles, and achieves the highest success on both subsets.

\paragraph{Controlled representation comparison.}
We further conduct a more rigorous controlled comparison by training all three formulations exclusively on our dedicated seven-task rotation-sensitive dataset, rather than simply partitioning the full-benchmark evaluation. The raw scalar, sine--cosine, and AngularFM models share the same backbone, demonstrations, training budget, and execution settings. Across three training seeds and three scene sets in Tab.~\ref{tab:wrist7}, sine--cosine improves success from 64.1\% to 71.2\%, demonstrating the benefit of continuous circular representations. AngularFM further improves success to 74.6\% through its manifold-preserving generation and
state-anchored source, achieving the highest mean across all three scene sets. These controlled results further support the benefits of aligning joint-angle representations and generative dynamics with circular geometry. The lower absolute success rates compared with the full RoboTwin 2.0 benchmark likely reflect the reduced scale and diversity of the seven-task training dataset, which limits the learned policies' overall manipulation capabilities. Appendix~\ref{sec:supp_offcircle} further analyzes how near-origin planar predictions amplify angular decoding sensitivity and reports norm-deviation diagnostics. By keeping generated states on the unit circle, AngularFM eliminates this radius-dependent amplification, providing a geometric guarantee distinct from prediction accuracy.

\subsection{Real-World Manipulation}
\label{sec:real_world}

On the six real-world tasks in Tab.~\ref{tab:real_world}, AngularFM succeeds in 47 of 60 trials (23 of 30 on the Franka and 24 of 30 on the AgileX), compared with 42 for $\pi_{0.5}$, 38 for raw scalar, and 32 for sine--cosine.
It is the best or tied-best method on every task except rack flipping, where $\pi_{0.5}$ succeeds in 7 trials and AngularFM in 6, and it is never below the raw scalar or sine--cosine model. More experimental results are provided in Appendix~\ref{sec:supp_real}.

\paragraph{Case study: block insertion.}
The Franka seats a cylindrical block upright in a round hole formed by two notched blocks, under three object layouts in Fig.~\ref{fig:real_deployment}. The cylinder enters only if it arrives aligned above the hole, so success depends on the final joint targets. The motion relies mainly on base rotation (J1) and the reach of J2 and J4 (Fig.~\ref{fig:teaser}), where an angular error is amplified by the length of the arm into a positional offset at the cylinder. Small errors in the generated angles can therefore make the cylinder miss the hole. AngularFM reaches each target through rotations on its joint circle that start near the measured joint angle, so its targets are not perturbed by the decoding of off-manifold samples. It succeeds in 8 of 10 trials, against 7 for sine--cosine and 6 for raw scalar.

\begin{table}[t]
\centering
\small
\caption{Success rates (\%) on rotation-sensitive manipulation tasks. Each scene-set entry is the mean
and sample standard deviation across three training seeds. Overall values
are computed across all nine training-seed--scene-set evaluations,
with 1,260 rollouts per method. Scene set 2 uses training-split scene seeds;
scene set 1 uses unseen seeds and scene set 3 a separately specified seed list.}
\label{tab:wrist7}
\setlength{\tabcolsep}{4mm}{} 
\resizebox{1.0\columnwidth}{!}{%
\begin{tabular}{l|ccc}
\toprule
Evaluation & Raw scalar & Sine--cosine & \textbf{AngularFM} \\
\midrule
Scene set 1 & $62.6\pm3.6$ & $70.5\pm3.7$ & $\mathbf{74.5}\pm1.5$ \\
Scene set 2 & $66.2\pm0.8$ & $72.4\pm2.5$ & $\mathbf{73.6}\pm4.7$ \\
Scene set 3 & $63.6\pm4.0$ & $70.5\pm4.8$ & $\mathbf{75.7}\pm1.2$ \\
\midrule
Overall Success & $64.1\pm3.1$ & $71.2\pm3.4$ & $\mathbf{74.6}\pm2.7$ \\
\bottomrule
\end{tabular}}
\end{table}

\begin{table}[t]
\centering
\small
\caption{Success counts on six real-world manipulation tasks,
with ten trials per task.}
\label{tab:real_world}
\resizebox{\columnwidth}{!}{%
\begin{tabular}{ll|cccc}
\toprule
\multirow{2}{*}{Robot}
& \multirow{2}{*}{Task}
& \multirow{2}{*}{$\pi_{0.5}$}
& \multicolumn{2}{c}{Fast-WAM-Joint}
& \multirow{2}{*}{\textbf{AngularFM}} \\
\cmidrule(lr){4-5}
& & & Raw scalar & Sine--cosine & \\
\midrule
\multirow{3}{*}{Franka} & Shoe placement  & 7/10 & 6/10 & 6/10 & \textbf{8/10} \\
 & Block insertion & 5/10 & 6/10 & 7/10 & \textbf{8/10} \\
 & Block stacking  & 6/10 & 4/10 & 3/10 & \textbf{7/10} \\
\midrule
\multirow{3}{*}{AgileX} & Cup inversion   & 8/10 & 8/10 & 5/10 & \textbf{9/10} \\
 & Plate insertion & \textbf{9/10} & 8/10 & \textbf{9/10} & \textbf{9/10} \\
 & Rack flipping   & \textbf{7/10} & 6/10 & 2/10 & 6/10 \\
\midrule
\multicolumn{2}{l|}{Overall Success} & 42/60 & 38/60 & 32/60 & \textbf{47/60} \\
\bottomrule
\end{tabular}%
}
\end{table}

\begin{table}[t]
\centering
\small
\caption{Source-distribution ablations on the rotation-sensitive task suite. CMAE is measured on validation windows.}
\label{tab:source_ablation}
\setlength{\tabcolsep}{4mm}{} 
\resizebox{\columnwidth}{!}{%
\begin{tabular}{l|c|cc}
\toprule
Source & $\kappa$ & Success (\%) $\uparrow$ & CMAE (rad) $\downarrow$ \\
\midrule
Uniform & --- & 60.7 & 0.0431 \\
State-anchored & 0.2 & 61.0 & 0.0410 \\
State-anchored & 0.5 & 66.0 & 0.0399 \\
State-anchored & 1 & 73.1 & \textbf{0.0368} \\
State-anchored & 2 & 75.2 & 0.0371 \\
State-anchored & 4 & \textbf{76.2} & 0.0372 \\
\bottomrule
\end{tabular}}
\end{table}

\subsection{Ablations and Analysis}
\label{sec:ablations}
Having examined the effects of joint-angle representations in Sec.~\ref{sec:rotation_sensitive}, we now investigate the state-anchored source distribution and its angular-noise precision on the rotation-sensitive task suite. Circular mean absolute error (CMAE) measures the shortest angular distance between predicted and demonstrated targets.

\paragraph{State-anchored source.}
With the architecture and rotation-based sampler unchanged, replacing a uniform source on $\mathbb S^1$ with the state-anchored source at $\kappa=0.5$ increases success from 60.7\% to 66.0\% in Tab.~\ref{tab:source_ablation}.
Validation CMAE decreases from 0.0431 to 0.0399 rad.
Both sources keep generation on the circle, so this 5.3-point gain isolates the benefit of anchoring generation at the current joint configuration.

\paragraph{Angular-noise precision.}
As shown in Tab.~\ref{tab:source_ablation}, increasing $\kappa$ from 1 to 2 and 4 improves success rates from 73.1\% to 75.2\% and 76.2\%, respectively, while validation CMAE remains nearly unchanged at 0.0368--0.0372 rad. These results suggest that concentrating the source around the current joint configuration benefits
closed-loop manipulation without substantially affecting angular prediction accuracy.
We therefore adopt $\kappa=4$ for the rotation-sensitive experiments.

\section{Conclusion}
\label{sec:conclusion}

We formulated three design criteria for joint-angle generation: continuous representations, consistency with circular geometry, and manifold-preserving generation.
They expose complementary limitations of raw scalar and sine--cosine representations under Euclidean flow matching.
A continuous representation alone proved insufficient, which supports aligning both the representation and the generation of joint angles with circular geometry.
AngularFM satisfies all three by generating joint-position targets on joint circles with an angular velocity field and a state-anchored source.
It outperforms both representations in controlled comparisons on our rotation-sensitive task suite and on two physical robots, and achieves strong results on LIBERO and RoboTwin 2.0.
\section*{LLM Usage Statement}
We used OpenAI ChatGPT to assist with language editing, manuscript organization, and refinement of technical explanations. The authors reviewed and verified all AI-assisted content and take full responsibility for the scientific claims and conclusions presented in this work.

{
    \small
    \bibliographystyle{ieeenat_fullname}
    \bibliography{main}

@String(CVPR= {IEEE Conf. Comput. Vis. Pattern Recog.})

@String(CVPR  = {CVPR})

@article{chi2025diffusion,
  title={Diffusion policy: Visuomotor policy learning via action diffusion},
  author={Chi, Cheng and Xu, Zhenjia and Feng, Siyuan and Cousineau, Eric and Du, Yilun and Burchfiel, Benjamin and Tedrake, Russ and Song, Shuran},
  journal={The International Journal of Robotics Research},
  volume={44},
  number={10-11},
  pages={1684--1704},
  year={2025},
  publisher={Sage Publications, Sage UK: London, England}
}

@article{intelligence2025pi_,
  title={$\pi_{0.5}$: a Vision-Language-Action Model with Open-World Generalization},
  author={Intelligence, Physical and Black, Kevin and Brown, Noah and Darpinian, James and Dhabalia, Karan and Driess, Danny and Esmail, Adnan and Equi, Michael and Finn, Chelsea and Fusai, Niccolo and others},
  journal={arXiv preprint arXiv:2504.16054},
  year={2025}
}

@article{li2024cogact,
  title={Cogact: A foundational vision-language-action model for synergizing cognition and action in robotic manipulation},
  author={Li, Qixiu and Liang, Yaobo and Wang, Zeyu and Luo, Lin and Chen, Xi and Liao, Mozheng and Wei, Fangyun and Deng, Yu and Xu, Sicheng and Zhang, Yizhong and others},
  journal={arXiv preprint arXiv:2411.19650},
  year={2024}
}

@article{brohan2023rt,
  title={Rt-2: Vision-language-action models transfer web knowledge to robotic control},
  author={Brohan, Anthony and Brown, Noah and Carbajal, Justice and Chebotar, Yevgen and Chen, Xi and Choromanski, Krzysztof and Ding, Tianli and Driess, Danny and Dubey, Avinava and Finn, Chelsea and others},
  journal={arXiv preprint arXiv:2307.15818},
  year={2023}
}

@article{ye2026world,
  title={World action models are zero-shot policies},
  author={Ye, Seonghyeon and Ge, Yunhao and Zheng, Kaiyuan and Gao, Shenyuan and Yu, Sihyun and Kurian, George and Indupuru, Suneel and Tan, You Liang and Zhu, Chuning and Xiang, Jiannan and others},
  journal={arXiv preprint arXiv:2602.15922},
  year={2026}
}

@article{zhu2025unified,
  title={Unified world models: Coupling video and action diffusion for pretraining on large robotic datasets},
  author={Zhu, Chuning and Yu, Raymond and Feng, Siyuan and Burchfiel, Benjamin and Shah, Paarth and Gupta, Abhishek},
  journal={arXiv preprint arXiv:2504.02792},
  year={2025}
}

@article{feng2026demystifying,
  title={Demystifying action space design for robotic manipulation policies},
  author={Feng, Yuchun and Zheng, Jinliang and Wang, Zhihao and Liu, Dongxiu and Li, Jianxiong and Pang, Jiangmiao and Wang, Tai and Zhan, Xianyuan},
  journal={arXiv preprint arXiv:2602.23408},
  year={2026}
}

@misc{universalrobots2024ur3e,
  author       = {{Universal Robots A/S}},
  title        = {{UR3e Technical Specification}},
  year         = {2024},
  month        = dec,
  howpublished = {\url{https://www.universal-robots.com/media/1807464/ur3e_e-series_datasheets_web.pdf}},
  note         = {Updated December 2024}
}

@article{chen2025robotwin,
  title={Robotwin 2.0: A scalable data generator and benchmark with strong domain randomization for robust bimanual robotic manipulation},
  author={Chen, Tianxing and Chen, Zanxin and Chen, Baijun and Cai, Zijian and Liu, Yibin and Li, Zixuan and Liang, Qiwei and Lin, Xianliang and Ge, Yiheng and Gu, Zhenyu and others},
  journal={arXiv preprint arXiv:2506.18088},
  year={2025}
}

@inproceedings{beyer2015biternion,
  title={Biternion nets: Continuous head pose regression from discrete training labels},
  author={Beyer, Lucas and Hermans, Alexander and Leibe, Bastian},
  booktitle={German Conference on Pattern Recognition},
  pages={157--168},
  year={2015},
  organization={Springer}
}

@inproceedings{chen2024flow,
  title={Flow matching on general geometries},
  author={Chen, Ricky TQ and Lipman, Yaron},
  booktitle={International Conference on Learning Representations},
  volume={2024},
  pages={47922--47945},
  year={2024}
}

@inproceedings{braun2024riemannian,
  title={Riemannian flow matching policy for robot motion learning},
  author={Braun, Max and Jaquier, No{\'e}mie and Rozo, Leonel and Asfour, Tamim},
  booktitle={2024 IEEE/RSJ International Conference on Intelligent Robots and Systems (IROS)},
  pages={5144--5151},
  year={2024},
  organization={IEEE}
}

@article{ding2025fast,
  title={Fast and robust visuomotor riemannian flow matching policy},
  author={Ding, Haoran and Jaquier, No{\'e}mie and Peters, Jan and Rozo, Leonel},
  journal={IEEE Transactions on robotics},
  year={2025},
  publisher={IEEE}
}

@article{liu2023libero,
  title={Libero: Benchmarking knowledge transfer for lifelong robot learning},
  author={Liu, Bo and Zhu, Yifeng and Gao, Chongkai and Feng, Yihao and Liu, Qiang and Zhu, Yuke and Stone, Peter},
  journal={Advances in Neural Information Processing Systems},
  volume={36},
  pages={44776--44791},
  year={2023}
}

@article{kim2024openvla,
  title={Openvla: An open-source vision-language-action model},
  author={Kim, Moo Jin and Pertsch, Karl and Karamcheti, Siddharth and Xiao, Ted and Balakrishna, Ashwin and Nair, Suraj and Rafailov, Rafael and Foster, Ethan and Lam, Grace and Sanketi, Pannag and others},
  journal={arXiv preprint arXiv:2406.09246},
  year={2024}
}

@article{black2024pi_0,
  title={$\pi_0$: A Vision-Language-Action Flow Model for General Robot Control},
  author={Black, Kevin and Brown, Noah and Driess, Danny and Esmail, Adnan and Equi, Michael and Finn, Chelsea and Fusai, Niccolo and Groom, Lachy and Hausman, Karol and Ichter, Brian and others},
  journal={arXiv preprint arXiv:2410.24164},
  year={2024}
}

@article{bjorck2025gr00t,
  title={Gr00t n1: An open foundation model for generalist humanoid robots},
  author={Bjorck, Johan and Casta{\~n}eda, Fernando and Cherniadev, Nikita and Da, Xingye and Ding, Runyu and Fan, Linxi and Fang, Yu and Fox, Dieter and Hu, Fengyuan and Huang, Spencer and others},
  journal={arXiv preprint arXiv:2503.14734},
  year={2025}
}

@article{hu2024video,
  title={Video prediction policy: A generalist robot policy with predictive visual representations},
  author={Hu, Yucheng and Guo, Yanjiang and Wang, Pengchao and Chen, Xiaoyu and Wang, Yen-Jen and Zhang, Jianke and Sreenath, Koushil and Lu, Chaochao and Chen, Jianyu},
  journal={arXiv preprint arXiv:2412.14803},
  year={2024}
}

@article{yuan2026fast,
  title={Fast-wam: Do world action models need test-time future imagination?},
  author={Yuan, Tianyuan and Dong, Zibin and Liu, Yicheng and Zhao, Hang},
  journal={arXiv preprint arXiv:2603.16666},
  year={2026}
}

@article{wan2025wan,
  title={Wan: Open and advanced large-scale video generative models},
  author={Wan, Team and Wang, Ang and Ai, Baole and Wen, Bin and Mao, Chaojie and Xie, Chen-Wei and Chen, Di and Yu, Feiwu and Zhao, Haiming and Yang, Jianxiao and others},
  journal={arXiv preprint arXiv:2503.20314},
  year={2025}
}

@inproceedings{jang2018qt,
  title={Scalable deep reinforcement learning for vision-based robotic manipulation},
  author={Kalashnikov, Dmitry and Irpan, Alex and Pastor, Peter and Ibarz, Julian and Herzog, Alexander and Jang, Eric and Quillen, Deirdre and Holly, Ethan and Kalakrishnan, Mrinal and Vanhoucke, Vincent and others},
  booktitle={Conference on robot learning},
  pages={651--673},
  year={2018},
  organization={PMLR}
}

@inproceedings{shridhar2023perceiver,
  title={Perceiver-actor: A multi-task transformer for robotic manipulation},
  author={Shridhar, Mohit and Manuelli, Lucas and Fox, Dieter},
  booktitle={Conference on Robot Learning},
  pages={785--799},
  year={2023},
  organization={PMLR}
}

@article{ke20243d,
  title={3d diffuser actor: Policy diffusion with 3d scene representations},
  author={Ke, Tsung-Wei and Gkanatsios, Nikolaos and Fragkiadaki, Katerina},
  journal={arXiv preprint arXiv:2402.10885},
  year={2024}
}

@inproceedings{zhou2019continuity,
  title={On the continuity of rotation representations in neural networks},
  author={Zhou, Yi and Barnes, Connelly and Lu, Jingwan and Yang, Jimei and Li, Hao},
  booktitle={2019 IEEE/CVF Conference on Computer Vision and Pattern Recognition (CVPR)},
  pages={5738--5746},
  year={2019},
  organization={IEEE}
}

@inproceedings{handa2020dexpilot,
  title={Dexpilot: Vision-based teleoperation of dexterous robotic hand-arm system},
  author={Handa, Ankur and Van Wyk, Karl and Yang, Wei and Liang, Jacky and Chao, Yu-Wei and Wan, Qian and Birchfield, Stan and Ratliff, Nathan and Fox, Dieter},
  booktitle={2020 IEEE international conference on robotics and automation (ICRA)},
  pages={9164--9170},
  year={2020},
  organization={IEEE}
}

@inproceedings{fu2024mobile,
  title={Mobile aloha: Learning bimanual mobile manipulation using low-cost whole-body teleoperation},
  author={Fu, Zipeng and Zhao, Tony Z and Finn, Chelsea},
  booktitle={8th Annual Conference on Robot Learning},
  year={2024}
}

@article{papamakarios2021normalizing,
  title={Normalizing flows for probabilistic modeling and inference},
  author={Papamakarios, George and Nalisnick, Eric and Rezende, Danilo Jimenez and Mohamed, Shakir and Lakshminarayanan, Balaji},
  journal={Journal of Machine Learning Research},
  volume={22},
  number={57},
  pages={1--64},
  year={2021}
}

@inproceedings{bi2026motus,
  title={Motus: A unified latent action world model},
  author={Bi, Hongzhe and Tan, Hengkai and Xie, Shenghao and Wang, Zeyuan and Huang, Shuhe and Liu, Haitian and Zhao, Ruowen and Feng, Yao and Xiang, Chendong and Rong, Yinze and others},
  booktitle={Proceedings of the IEEE/CVF Conference on Computer Vision and Pattern Recognition},
  pages={35101--35113},
  year={2026}
}

@article{li2026causal,
  title={Causal world modeling for robot control},
  author={Li, Lin and Zhang, Qihang and Luo, Yiming and Yang, Shuai and Wang, Ruilin and Han, Fei and Yu, Mingrui and Gao, Zelin and Xue, Nan and Zhu, Xing and others},
  journal={arXiv preprint arXiv:2601.21998},
  year={2026}
}

@inproceedings{zheng2026x,
  title={X-vla: Soft-prompted transformer as scalable cross-embodiment vision-language-action model},
  author={Zheng, Jinliang and Li, Jianxiong and Wang, Zhihao and Liu, Dongxiu and Kang, Xirui and Feng, Yuchun and Zheng, Yinan and Zou, Jiayin and Chen, Yilun and Zeng, Jia and others},
  booktitle={International Conference on Learning Representations},
  volume={2026},
  pages={60580--60606},
  year={2026}
}

@article{lipman2022flow,
  title={Flow matching for generative modeling},
  author={Lipman, Yaron and Chen, Ricky TQ and Ben-Hamu, Heli and Nickel, Maximilian and Le, Matt},
  journal={arXiv preprint arXiv:2210.02747},
  year={2022}
}

@article{bu2505learning,
  title={{UniVLA}: Learning to Act Anywhere with Task-centric Latent Actions},
  author={Bu, Qingwen and Yang, Yanting and Cai, Jisong and Gao, Shenyuan and Ren, Guanghui and Yao, Maoqing and Luo, Ping and Li, Hongyang},
  journal={arXiv preprint arXiv:2505.06111},
  year={2025}
}

@article{li2026cogvla,
  title={CogVLA: Cognition-aligned vision-language-action models via instruction-driven routing \& sparsification},
  author={Li, Wei and Zhang, Renshan and Shao, Rui and He, Jie and Nie, Liqiang},
  journal={Advances in neural information processing systems},
  volume={38},
  pages={137646--137675},
  year={2026}
}

@article{zhang2026dreamvla,
  title={Dreamvla: a vision-language-action model dreamed with comprehensive world knowledge},
  author={Zhang, Wenyao and Liu, Hongsi and Qi, Zekun and Wang, Yunnan and Yu, Xinqiang and Zhang, Jiazhao and Dong, Runpei and He, Jiawei and Wang, He and Zhang, Zhizheng and others},
  journal={Advances in Neural Information Processing Systems},
  volume={38},
  pages={24195--24228},
  year={2026}
}

@inproceedings{zhao2025cot,
  title={Cot-vla: Visual chain-of-thought reasoning for vision-language-action models},
  author={Zhao, Qingqing and Lu, Yao and Kim, Moo Jin and Fu, Zipeng and Zhang, Zhuoyang and Wu, Yecheng and Li, Zhaoshuo and Ma, Qianli and Han, Song and Finn, Chelsea and others},
  booktitle={2025 IEEE/CVF Conference on Computer Vision and Pattern Recognition (CVPR)},
  pages={1702--1713},
  year={2025},
  organization={IEEE}
}

@article{chen2026abot,
  title={Abot-m0. 5: Unified mobility-and-manipulation world action model},
  author={Chen, Ronghan and Yang, Yandan and Tang, Zuojin and Huo, Dongjie and Lin, Tong and Wu, Haoning and Liu, Haoyun and Chen, Yuzhi and Zheng, Lulu and Yuan, Botai and others},
  journal={arXiv preprint arXiv:2607.00678},
  year={2026}
}

@article{wang2026qwen,
  title={Qwen-vla: Unifying vision-language-action modeling across tasks, environments, and robot embodiments},
  author={Wang, Qiuyue and Li, Mingsheng and Guan, Jian and Ye, Jinhui and Xie, Sicheng and Liu, Yitao and Chen, Junhao and Liang, Zhixuan and Zhang, Jie and Hu, Xintong and others},
  journal={arXiv preprint arXiv:2605.30280},
  year={2026}
}

@article{yuan2026qwen,
  title={Qwen-robotmanip technical report: Alignment unlocks scale for robotic manipulation foundation models},
  author={Yuan, Haoqi and Liang, Zhixuan and Chen, Anzhe and Wang, Ye and Li, Haoyang and Lin, Pei and Huang, Yiyang and Lei, Zixing and Zhang, Tong and Zhang, Jiazhao and others},
  journal={arXiv preprint arXiv:2606.17846},
  year={2026}
}

@article{zhang2026imagewam,
  title={ImageWAM: Do World Action Models Really Need Video Generation, or Just Image Editing?},
  author={Zhang, Yuyang and Zhang, Wenyao and Qi, Zekun and Zhang, He and Lin, Haitao and Zhang, Jingbo and Mu, Yao and Yang, Xiaokang and Zeng, Wenjun and Jin, Xin},
  journal={arXiv preprint arXiv:2606.19531},
  year={2026}
}

@article{yang2026abot,
  title={Abot-m0: Vla foundation model for robotic manipulation with action manifold learning},
  author={Yang, Yandan and Zeng, Shuang and Lin, Tong and Chang, Xinyuan and Qi, Dekang and Xiao, Junjin and Liu, Haoyun and Chen, Ronghan and Chen, Yuzhi and Huo, Dongjie and others},
  journal={arXiv preprint arXiv:2602.11236},
  year={2026}
}

@article{cai2026aha,
  title={AHA-WAM: Asynchronous Horizon-Adaptive World-Action Modeling with Observation-Guided Context Routing},
  author={Cai, Jisong and Ling, Long and Chu, Shiwei and Liu, Zhongshan and Kang, Jiayue and Liang, Zhixuan and Xu, Wenjie and Mao, Yinan and Zhang, Weinan and Yang, Xiaokang and others},
  journal={arXiv preprint arXiv:2606.09811},
  year={2026}
}

@article{gong2026fast,
  title={Fast-VLA: You Only Think Once for Robot Manipulation},
  author={Gong, Kehong and Hou, Guanli and Zheng, Xiaoxu and Wang, Qi and Xu, Mingxi and Zhai, Anbang and He, Xiaoyu and Zhang, Mingyuan and Mi, Michael Bi},
  journal={arXiv preprint arXiv:2610.21389},
  year={2026}
}

@inproceedings{wu2024gello,
  title={Gello: A general, low-cost, and intuitive teleoperation framework for robot manipulators},
  author={Wu, Philipp and Shentu, Yide and Yi, Zhongke and Lin, Xingyu and Abbeel, Pieter},
  booktitle={2024 IEEE/RSJ International Conference on Intelligent Robots and Systems (IROS)},
  pages={12156--12163},
  year={2024},
  organization={IEEE}
}

@article{akbari2026flash,
  title={Flash-WAM: Modality-Aware Distillation for World Action Models},
  author={Akbari, Arman and Zhang, Ci and Akbari, Arash and Zhao, Lin and Chen, Yixiao and Chen, Weiwei and Zhang, Xuan and Yuan, Geng and Wang, Yanzhi},
  journal={arXiv preprint arXiv:2606.05254},
  year={2026}
}
}

%
%
%
%

\clearpage
\setcounter{page}{1}
\maketitlesupplementary
\appendix

\setcounter{table}{0}\renewcommand{\thetable}{S\arabic{table}}
\setcounter{figure}{0}\renewcommand{\thefigure}{S\arabic{figure}}
\setcounter{equation}{0}\renewcommand{\theequation}{S\arabic{equation}}
\providecommand{\theHtable}{}\renewcommand{\theHtable}{supp.\arabic{table}}
\providecommand{\theHfigure}{}\renewcommand{\theHfigure}{supp.\arabic{figure}}
\providecommand{\theHequation}{}\renewcommand{\theHequation}{supp.\arabic{equation}}

\newcommand{\mainsec}[1]{Sec.~#1 of the main paper}
\newcommand{\maintab}[1]{Tab.~#1 of the main paper}
\newcommand{\mainfig}[1]{Fig.~#1 of the main paper}
\newcommand{\maineq}[1]{Eq.~(#1) of the main paper}
\newcommand{\mainalg}[1]{Alg.~#1 of the main paper}

\newcounter{suppprop}
\newcommand{\suppprop}[1]{\par\smallskip\noindent\refstepcounter{suppprop}\textbf{Proposition~\thesuppprop} (#1)\textbf{.}\ }
\newcommand{\suppproof}{\par\smallskip\noindent\textit{Proof.}\ }
\newcommand{\suppqed}{\hfill$\square$\par\smallskip}
\newcommand{\supptask}[1]{\texttt{#1}}
\newcommand{\suppgraphic}[2][width=\linewidth]{%
  \IfFileExists{#2}{\includegraphics[#1]{#2}}{%
    \fbox{\parbox[c][0.2\textheight][c]{0.96\linewidth}{\centering\small
      Figure file \texttt{\detokenize{#2}} not found.}}}}

\noindent
This supplementary material provides the details that the main paper refers to.
\cref{sec:supp_suite} describes the rotation-sensitive task suite, including task selection, data collection, and the training and evaluation protocols.
\cref{sec:supp_impl} gives the training and implementation details of all experiments.
\cref{sec:supp_offcircle} analyzes how off-circle planar predictions affect angular decoding and reports norm-deviation diagnostics.
\cref{sec:supp_derivations} collects derivations that complement the method section.
\cref{sec:supp_real} presents additional details and qualitative results of the real-world experiments.

\section{Rotation-Sensitive Task Suite}
\label{sec:supp_suite}

This section details the suite of seven rotation-sensitive tasks introduced in \mainsec{4.1}.

\subsection{Task Selection}
\label{sec:supp_selection}

We select the tasks among the 50 tasks of RoboTwin~2.0~\cite{chen2025robotwin} according to the angular variation of the wrist-roll joints, which are the last revolute joints of the two arms, in the released demonstrations.
The suite consists of the seven tasks \supptask{rotate\_qrcode}, \supptask{place\_dual\_shoes}, \supptask{move\_stapler\_pad}, \supptask{place\_shoe}, \supptask{place\_mouse\_pad}, \supptask{press\_stapler}, and \supptask{turn\_switch}.
Their demonstrations contain large wrist rotations, and the same wrist orientation is demonstrated with joint values that differ by a full turn.
We refer to these two alternatives as the short and the long branch.
Policies trained on these tasks are therefore exposed to angular wraparound.
The selection was fixed before any of the compared policies was trained.

\subsection{Data Collection}
\label{sec:supp_collection}

In the released demonstrations, the two branches are a by-product of motion planning, and their proportion cannot be controlled.
We therefore collect new demonstrations for the seven tasks with the scripted experts of RoboTwin~2.0 and turn the branch into a controlled factor.
After each successful motion plan, we rewrite the wrist-roll trajectory so that the episode ends on a requested branch.
Let $q_T$ be the planned final wrist-roll angle, measured from the home configuration, and let $\operatorname{wrap}(\cdot)$ map an angle to $[-\pi,\pi)$.
The short branch ends at $q_{\mathrm{short}}=\operatorname{wrap}(q_T)$.
The long branch ends at
\begin{equation}
q_{\mathrm{long}}=q_{\mathrm{short}}-\operatorname{sign}(q_{\mathrm{short}})\,2\pi ,
\label{eq:supp_long_branch}
\end{equation}
which is the same wrist orientation reached through the complementary arc.
The rewrite is skipped when the long branch would violate the joint limit of $\pm10$\,rad.

For every task, we collect 100 successful episodes for each of the two scene conditions of RoboTwin~2.0, clean and randomized, and for each of the two requested branches.
The randomized condition applies the domain randomization of RoboTwin~2.0 to backgrounds, clutter objects, table height, and lighting.
This gives $7\times2\times2\times100=2{,}800$ episodes with 466,119 frames at a control period of 0.06\,s.
The episodes last 10.0\,s on average and 7.77 hours in total.
We label each episode by its measured net wrist rotation, as long if it exceeds $\pi$ and as short otherwise.
The suite contains 862 long and 1,938 short episodes, because a requested long branch is not always feasible.
Joint angles are stored without wrapping and reach 6.31\,rad.
The largest wrist rotation between consecutive frames is 0.295\,rad.

We split the episodes by hashing the task, the scene condition, and the scene seed.
The training, validation, and test subsets contain 2,256, 283, and 261 episodes.
The 1,756 distinct scenes are thus disjoint across the subsets, and the short and long variants of a scene always fall into the same subset.
During training, the two branches are sampled with equal probability.

\subsection{Training Protocol}
\label{sec:supp_suite_training}

The raw scalar, sine--cosine, and AngularFM policies differ only in the action representation and its generative process (\cref{sec:supp_baselines}).
They share the backbone, the demonstrations, the training schedule, and the execution settings, and the proprioceptive input uses the sine--cosine encoding in all three.
AngularFM uses $\kappa=4$.
We repeat training with three random seeds and evaluate the final checkpoint of each run.

\subsection{Evaluation Protocol}
\label{sec:supp_suite_eval}

Policies are evaluated in closed loop in RoboTwin~2.0 on the bimanual AgileX robot, with $N=10$ sampling steps and unseen language instructions.
Each predicted chunk contains $H=32$ joint-position targets, and the first 24 are executed before the next chunk is predicted.

A scene set consists of 10 scenes for each task and scene condition, which gives 140 episodes.
Scene set~1 uses scene seeds that do not appear in the training subset.
Scene set~2 uses scene seeds of the training subset.
Scene set~3 uses a separately specified list of unseen seeds.
All policies are evaluated on identical scenes, which allows paired comparisons.
Every training seed is evaluated on every scene set.
This yields nine evaluations and 1,260 rollouts per method, which are summarized in \maintab{3}.
Paired differences between two formulations are computed over pairs of episodes that share the training seed and the scene, with 95\% bootstrap intervals over episodes.

\subsection{Statistics of the Executed Motions}
\label{sec:supp_suite_results}

The success rates of the three formulations are reported in \maintab{3}.
To relate them to the wraparound discontinuity of the raw scalar representation, we analyze the joint commands that were executed in the same rollouts, with 1,260 episodes per method (\cref{tab:supp_motion}).
A $2\pi$ flip is an executed step at which the command of a revolute joint changes by more than $\pi$.
The remaining statistics are computed after wrapping the differences of consecutive commands to $[-\pi,\pi)$, and the third row of the table counts the steps at which a wrapped difference still exceeds $\pi/2$.
The boundary discontinuity is the absolute difference between the last executed command of a chunk and the first command of the next chunk, averaged over the 12 revolute joints.
The steps counted in the third row are excluded from it.
The peak acceleration is the largest second-order difference of the commands within an episode, divided by the squared control period.

Raw scalar is the only formulation that produces $2\pi$ flips.
They occur at 2,630 steps and in 583 of its 1,260 episodes, whereas the two circle-based formulations produce none.
The boundary discontinuity of the circle-based formulations is about 30\% smaller than that of raw scalar, and their peak acceleration is less than half.
Between the two, AngularFM has the smaller boundary discontinuity, with a paired difference of $-0.77\times10^{-3}$\,rad and an interval of $[-1.19,-0.32]\times10^{-3}$\,rad.
Its peak acceleration is higher than that of sine--cosine by 11.4\,rad/s$^2$, with an interval of $[8.1,14.9]$\,rad/s$^2$.

\begin{table}[t]
\centering
\small
\caption{Motion statistics of the executed joint commands on the rotation-sensitive suite, with 1,260 episodes per method. The first three rows are counts. The last two rows are means and standard deviations over the nine combinations of training seed and scene set.}
\label{tab:supp_motion}
\setlength{\tabcolsep}{3.5pt}
\resizebox{\columnwidth}{!}{%
\begin{tabular}{l|ccc}
\toprule
 & Raw scalar & Sine--cosine & AngularFM \\
\midrule
Steps with a $2\pi$ flip            & 2,630 & 0 & 0 \\
Episodes with a $2\pi$ flip         & 583   & 0 & 0 \\
Steps with a change $>\pi/2$        & 167   & 8 & 19 \\
Boundary disc.\ ($10^{-3}$\,rad)    & $25.6\pm1.4$ & $18.2\pm1.1$ & $17.4\pm0.4$ \\
Peak acc.\ (rad/s$^2$)              & $103\pm8$    & $37\pm3$     & $48\pm4$ \\
\bottomrule
\end{tabular}}
\end{table}

\section{Training and Implementation Details}
\label{sec:supp_impl}

This section complements the implementation details of \mainsec{4.1}.

\subsection{Architecture and Action Interface}
\label{sec:supp_arch}

AngularFM adopts the model architecture of Fast-WAM-Joint~\cite{yuan2026fast}, in which the action tokens attend to the video tokens.
At inference, the video latents and the actions are denoised jointly.
The video expert is the diffusion transformer of Wan2.2-5B~\cite{wan2025wan} with 30 layers and a hidden dimension of 3072.
The action expert has 30 layers with a hidden dimension of 1024.
The video VAE and the text encoder are frozen, and all other parameters, 6.0B in total, are trained.

For a robot with $n$ revolute joints and $g$ gripper dimensions, the action expert reads $2n+g$ values per future action step.
They are the sine--cosine encoding of the $n$ noisy phases and the $g$ noisy gripper values, normalized to zero mean and unit variance.
The action expert predicts $n+g$ values per step, namely $n$ angular velocities and $g$ gripper velocities.
A linear layer maps the input to the hidden dimension, and a linear head produces the output.
The current joint positions are encoded in the same $2n+g$ layout, mapped by a linear layer to a single token, and appended to the text tokens.
These three linear layers form the action interface.
They are the only layers whose size depends on the joint layout and the action representation.
For the bimanual AgileX robot of RoboTwin~2.0, $n=12$ and $g=2$, so the input has 26 dimensions and the output has 14.

Grippers follow Euclidean flow matching with a standard Gaussian source and linear paths.
The action loss averages the squared velocity errors over the $n+g$ dimensions and is added to the video flow matching loss with equal weight.
Each training sample contains $H=32$ actions and nine video frames, one for every four action steps.
The camera views are arranged on a single canvas before VAE encoding.
In RoboTwin~2.0, the head view at $256\times320$ is placed above the two wrist views at $128\times160$ each, which gives a $384\times320$ image.

\subsection{Flow Time, Loss Weight, and Sampling Grid}
\label{sec:supp_time}

The flow time is sampled as
\begin{equation}
s=\frac{1-u}{1+4u},\qquad u\sim\mathcal U(0,1),
\label{eq:supp_time}
\end{equation}
independently for the video stream and the action stream.
This distribution $p$ places more samples near the source, with a median flow time of $1/6$.
The loss weight is
\begin{equation}
w(s)=\frac{1}{c}\left[\exp\!\left(-2\left(s-\tfrac12\right)^2\right)-e^{-1/2}\right],
\label{eq:supp_weight}
\end{equation}
where $c\approx0.190$ gives $w$ a unit mean under $p$.
The weight is positive for $0<s<1$ and peaks at $s=1/2$ with a value of 2.08.
Sampling uses $N=10$ steps on the grid
\begin{equation}
s_k=\frac{k}{5N-4k},\qquad k=0,\ldots,N,
\label{eq:supp_grid}
\end{equation}
which is shared by the video and action streams, with a classifier-free guidance scale of 1.0.
The joint phases are sampled and integrated in single precision, while the network runs in bfloat16 precision.

\subsection{Experiment-Specific Settings}
\label{sec:supp_settings}

\cref{tab:supp_settings} summarizes the settings that differ between the experiments.
All experiments use AdamW with $\beta_1=0.9$, $\beta_2=0.95$, a weight decay of 0.01, and gradient clipping at 1.0.
In every training stage, the learning rate follows a linear warm-up over the first 10\% of the steps and a cosine decay to 1\% of its peak.
Simulation training uses neither image augmentation nor an exponential moving average of the weights.

\begin{table*}[t]
\centering
\small
\caption{Experiment-specific settings of AngularFM.}
\label{tab:supp_settings}
\setlength{\tabcolsep}{4.5pt}
\begin{tabular}{l|ccc|cc}
\toprule
 & & Rotation-sensitive & & \multicolumn{2}{c}{Real robots} \\
 & RoboTwin 2.0 & suite & LIBERO & Franka & AgileX \\
\midrule
Revolute joints $n$   & 12 & 12 & 7 & 7 & 12 \\
Input / output dim.   & 26 / 14 & 26 / 14 & 15 / 8 & 15 / 8 & 26 / 14 \\
Precision $\kappa$    & 2 & 4 & 2 & 2 & 2 \\
Camera views          & 3 & 3 & 2 & 2 & 3 \\
Image canvas          & $384\times320$ & $384\times320$ & $480\times320$ & $224\times448$ & $384\times320$ \\
Training seeds        & 42 & 42, 43, 44 & 42 & 42 & 42 \\
\bottomrule
\end{tabular}
\end{table*}

\paragraph{RoboTwin 2.0.}
We use the preprocessed RoboTwin~2.0 dataset released with Fast-WAM~\cite{yuan2026fast}, hold out 900 demonstrations for validation, and train on the remaining 26,600.

\paragraph{Rotation-sensitive suite.}
The settings are given in \cref{sec:supp_suite_training}.
The demonstrations are processed in the same way as for RoboTwin~2.0.

\paragraph{LIBERO.}
We derive joint-position actions from the LIBERO demonstrations~\cite{liu2023libero}.
The state consists of the seven arm joint angles and the gripper width, and the action consists of the joint angles of the next step and the gripper command.
The seven arm joints are represented as phases, so that $n=7$ and $g=1$.
The two camera views are stacked into a $480\times320$ image.
At test time, the predicted joint-position targets are tracked by a joint-space position controller.

\paragraph{Real robots.}
The settings are given in \cref{sec:supp_real}.

\subsection{Baselines}
\label{sec:supp_baselines}

\paragraph{Sine--cosine.}
The action expert reads and predicts $2n+g$ values, and Euclidean flow matching with a standard Gaussian source is applied to all of them.
The two squared errors of a sine--cosine pair are summed and counted as one dimension, so that the loss averages over the same $n+g$ dimensions as in AngularFM.
After sampling, each pair is decoded with $\operatorname{atan2}$, and the joint-position targets are reconstructed with the same sequential phase lifting as in AngularFM.

\paragraph{Raw scalar.}
The $n+g$ joint values are normalized to zero mean and unit variance and generated with Euclidean flow matching.
The outputs are clipped to the joint limits and executed directly.
The proprioceptive input uses the same sine--cosine encoding as in the other two formulations.

\paragraph{Shared settings.}
On the rotation-sensitive suite, the three formulations share the demonstrations, the training schedule, and the execution settings (\cref{sec:supp_suite_training,sec:supp_suite_eval}).

\subsection{Evaluation on RoboTwin 2.0}
\label{sec:supp_robotwin_eval}

We use the evaluator of RoboTwin~2.0 with unseen language instructions, $N=10$ sampling steps, and 24 executed actions per predicted chunk.
The results of the other methods in \maintab{2} are reported values from the literature.
The task-subset analysis in \mainfig{2} groups the 50 tasks into the seven rotation-sensitive tasks of \cref{sec:supp_selection} and the 43 complementary tasks.
The values of Fast-WAM and Fast-WAM-Joint in that figure are computed from the per-task results reported in~\cite{yuan2026fast}, as the average of the clean and randomized conditions.

\section{Analysis of Off-Circle Predictions}
\label{sec:supp_offcircle}

This section supports the discussions in Sec.~3.1 and Sec.~4.3 of the main paper.
We analyze the sensitivity of angular decoding to the norm of a planar prediction, explain why planar flow matching produces states inside the unit circle, and report norm-deviation diagnostics of the sine--cosine baseline.
We use the notation of \mainsec{3}, with $\mathbf u(q)=(\cos q,\sin q)^\top$ and the rotation $K$ by $\pi/2$.

\subsection{Radial and Tangential Dynamics}
\label{sec:supp_polar}

Consider one joint and one future action step.
Write a planar state as $\mathbf x=r\,\mathbf u(\phi)$ with radius $r>0$ and phase $\phi$, and decompose its velocity into a radial and a tangential part, $\dot{\mathbf x}=\alpha\,\mathbf u(\phi)+\beta\,K\mathbf u(\phi)$.
Differentiating $\mathbf x=r\,\mathbf u(\phi)$ gives $\dot{\mathbf x}=\dot r\,\mathbf u(\phi)+r\dot\phi\,K\mathbf u(\phi)$, and therefore
\begin{equation}
\dot r=\alpha,\qquad \dot\phi=\frac{\beta}{r}.
\label{eq:supp_polar}
\end{equation}
The radial part changes only the radius, and the tangential part changes only the phase, with a gain of $1/r$.
A state that starts on the circle stays on it if and only if $\alpha=0$ along the whole trajectory, which leaves the angular velocity as the only degree of freedom, as in \maineq{3}.

\subsection{Sensitivity of Angular Decoding}
\label{sec:supp_sensitivity}

The sine--cosine baseline decodes a generated planar state $\hat{\mathbf x}=r\,\mathbf u(\hat q)$ into the angle $\hat q=\operatorname{atan2}(\hat x_2,\hat x_1)$.
Let a perturbation $\mathbf e$ of this state have the radial component $e_\parallel=\mathbf u(\hat q)^\top\mathbf e$ and the tangential component $e_\perp=(K\mathbf u(\hat q))^\top\mathbf e$.
The decoded angle changes by
\begin{equation}
\delta\hat q=\operatorname{atan2}\!\left(e_\perp,\;r+e_\parallel\right)\approx\frac{e_\perp}{r}.
\label{eq:supp_gain}
\end{equation}
A tangential error of a given size therefore causes an angular error that grows as the state approaches the origin.
Compared with a state on the circle, it is doubled at $r=0.5$ and multiplied by ten at $r=0.1$, and the decoded angle is undefined at the origin.

A complementary bound follows from the distance to the target.
Let $\mathbf z_1=\mathbf u(q_1)$ be the target phase and $\varepsilon=\|\hat{\mathbf x}-\mathbf z_1\|_2$ the planar error of the generated state.
For $\varepsilon<1$, the disk of radius $\varepsilon$ around $\mathbf z_1$ is seen from the origin under the half-angle $\arcsin\varepsilon$, so that
\begin{equation}
\left|\operatorname{ang}\!\left(\mathbf z_1,\mathbf u(\hat q)\right)\right|\le\arcsin\varepsilon ,
\label{eq:supp_arcsin}
\end{equation}
and the bound is attained.
For $\varepsilon\ge1$, the disk contains the origin and the decoded angle can be arbitrarily wrong.
On the circle, in contrast, an error of arc length $\varepsilon$ is an angular error of exactly $\varepsilon$.

\subsection{Why Planar States Leave the Circle}
\label{sec:supp_why}

\paragraph{Linear paths cross the disk.}
The linear path between two points on the circle with phase difference $\Delta$ has the radius
\begin{equation}
r(s)=\sqrt{1-2s(1-s)(1-\cos\Delta)} ,
\label{eq:supp_chord}
\end{equation}
with the minimum $|\cos(\Delta/2)|$ at $s=1/2$.
The Gaussian source of \maineq{1} adds to this effect.
For $\mathbf x_0\sim\mathcal N(0,I_2)$ and a uniformly distributed target phase, a Monte Carlo evaluation with $3\times10^6$ paths gives a median minimum radius of 0.63 along the path.
The path comes closer than 0.5 to the origin with probability 0.40, closer than 0.3 with probability 0.23, and closer than 0.1 with probability 0.07.
The training states thus cover the interior of the disk, including the region where the gain in \cref{eq:supp_gain} is large.

\paragraph{Finite-step sampling returns a conditional mean.}
On the linear path $\mathbf X_s=(1-s)\mathbf X_0+s\mathbf Z_1$ of the full chunk, the conditional velocity can be written as $\mathbf Z_1-\mathbf X_0=(\mathbf Z_1-\mathbf X_s)/(1-s)$.
The minimizer of the Euclidean objective in \maineq{2} is its conditional expectation,
\begin{equation}
\mathbf V^\star(\mathbf X,s,o)=\frac{\mathbf M(\mathbf X,s,o)-\mathbf X}{1-s},
\label{eq:supp_vstar}
\end{equation}
where $\mathbf M(\mathbf X,s,o)=\mathbb E\left[\mathbf Z_1\mid\mathbf X_s=\mathbf X,o\right]$ is the posterior mean of the target phases.
Let $\mathbf X'$ denote the state of the sampler before its last Euler step, which leads from $s'=s_{N-1}$ to $s_N=1$.
With the exact field $\mathbf V^\star$, this step returns
\begin{equation}
\widehat{\mathbf X}_1=\mathbf X'+(1-s')\,\mathbf V^\star(\mathbf X',s',o)=\mathbf M(\mathbf X',s',o).
\label{eq:supp_mean}
\end{equation}
Every sine--cosine pair of this output is a mean of unit vectors.
Such a mean has a norm of at most one, with equality only if the posterior of the corresponding phase is concentrated on a single value.
Its norm is the mean resultant length of the posterior.
For a wrapped normal posterior with standard deviation $\sigma$, this length is $e^{-\sigma^2/2}$, so that a norm of 0.98 corresponds to $\sigma\approx0.2$\,rad and a norm of 0.8 to $\sigma\approx0.67$\,rad.
For two equally likely phases that are $\Delta$ apart, it is $|\cos(\Delta/2)|$.
Even with an exact velocity field, a sampler with a finite number of steps thus ends inside the circle whenever the target phase is not yet determined at the last step, and the distance to the circle measures the remaining uncertainty.
With the grid of \cref{eq:supp_grid} and $N=10$, the last step starts at $s_{N-1}=0.64$.
The uncertain predictions are also the ones for which decoding is most sensitive according to \cref{eq:supp_gain}.

\subsection{Norm-Deviation Diagnostics}
\label{sec:supp_diagnostics}

A generated chunk contains one sine--cosine pair $\widehat{\mathbf x}_1$ for every joint and future action step.
We measure the norm deviation $\big|\|\widehat{\mathbf x}_1\|_2-1\big|$ of these pairs for a sine--cosine policy trained on the rotation-sensitive suite for 10 epochs.
For AngularFM, the same quantity is zero for every source state, intermediate state, and generated endpoint.

\paragraph{Open loop.}
On the 18,159 windows of the test subset, the norm deviation averaged over the $32\times12$ pairs of a window has a mean of 0.021, a 95th percentile of 0.031, and a maximum of 0.309.
The largest deviation within a window has a mean of 0.131, a 95th percentile of 0.259, and a 99th percentile of 0.606.
At least one pair deviates by more than 0.2 in 10.8\% of the windows and by more than 0.5 in 1.2\%.
The deviation shrinks along the sampling trajectory, from 0.104 in its early part to 0.030 in its middle part and 0.021 at its end.
Increasing the number of sampling steps does not remove it, with an average deviation of 0.025 for $N=4$, 0.021 for $N=10$, and 0.020 for $N=32$.
Over training, it decreases from 0.060 after 1k steps to 0.021 at the end.

The deviation is related to the angular error.
Across windows, the average norm deviation and the circular mean absolute error have a Spearman correlation of 0.56.
The 10\% of the windows with the largest maximum deviation have an error of 0.078\,rad, compared with 0.035\,rad for the remaining windows.
\cref{tab:supp_pairs} resolves this relation at the level of single pairs.
Pairs with a norm between 0.95 and 1.05 are decoded with a mean error of 0.035\,rad.
The error is twelve times larger for norms between 0.8 and 0.95 and exceeds 1\,rad for norms below 0.8.
Such pairs are rare and occur in the windows with wrist rotation. No pair has a norm below 0.8 in the windows without it. However, their effect in closed loop need not be small: every predicted chunk is executed for 24 steps, and the closed-loop logs below show that episodes containing a large deviation succeed less often.

\begin{table}[t]
\centering
\small
\caption{Angular error of the decoded sine--cosine pairs as a function of their norm, in open loop on 2,270 held-out windows of the rotation-sensitive suite, with $32\times12$ pairs per window.}
\label{tab:supp_pairs}
\setlength{\tabcolsep}{5pt}
\begin{tabular}{l|rrc}
\toprule
Norm of the pair & Pairs & Share (\%) & Mean error (rad) \\
\midrule
below 0.5   & 156     & 0.018  & 1.484 \\
0.5--0.8    & 257     & 0.029  & 1.079 \\
0.8--0.95   & 5,717   & 0.656  & 0.422 \\
0.95--1.05  & 862,211 & 98.914 & 0.035 \\
1.05--1.2   & 3,334   & 0.382  & 0.452 \\
above 1.2   & 5       & 0.001  & 1.609 \\
\bottomrule
\end{tabular}
\end{table}

\paragraph{Closed loop.}
We log the pairs generated by the same policy during closed-loop evaluation on the three scene sets of \cref{sec:supp_suite_eval}.
\cref{tab:supp_norms} shows the distribution of their norms on scene set~3.
About half of the pairs lie within 0.01 of the unit circle.
The larger deviations point inward, as predicted by \cref{eq:supp_mean}: 4.5\% of the pairs have a norm below 0.95, whereas 0.9\% have a norm above 1.05.
Norms above one cannot arise from \cref{eq:supp_mean} and reflect the approximation error of the learned field.
On scene sets~1 and~2, the share of pairs with a norm below 0.95 is 9.1\% and 2.6\%, where only a part of the episodes of scene set~1 was logged.
The value of scene set~2 is obtained on scene seeds of the training subset, which indicates that off-circle predictions are not limited to unseen scenes.

Off-circle predictions accompany failures.
On the three scene sets, the average norm deviation is 0.020, 0.016, and 0.020 in failed episodes, compared with 0.013 to 0.014 in successful episodes.
Episodes in which at least one pair deviates by more than 0.5 succeed less often than the remaining episodes, with success rates of 54.5\%, 38.5\%, and 60.0\% compared with 70.5\%, 80.3\%, and 70.8\%.
These episodes are few, between 11 and 20 of the 140 episodes of a scene set.
The diagnostics cannot separate the two mechanisms of \cref{sec:supp_why,sec:supp_sensitivity}: an uncertain prediction reduces the norm, and a reduced norm amplifies the remaining error.

\begin{table}[t]
\centering
\small
\caption{Norms of the sine--cosine pairs generated in closed loop on scene set~3, with 516,480 pairs from 140 episodes.}
\label{tab:supp_norms}
\setlength{\tabcolsep}{5pt}
\begin{tabular}{l|rr}
\toprule
Norm of the pair & Pairs & Share (\%) \\
\midrule
below 0.5    & 149     & 0.03  \\
0.5--0.8     & 2,322   & 0.45  \\
0.8--0.95    & 20,839  & 4.03  \\
0.95--0.99   & 143,730 & 27.83 \\
0.99--1.01   & 250,537 & 48.51 \\
1.01--1.05   & 94,359  & 18.27 \\
1.05--1.2    & 4,529   & 0.88  \\
above 1.2    & 15      & 0.003 \\
\bottomrule
\end{tabular}
\end{table}

\subsection{Guarantee of AngularFM}
\label{sec:supp_guarantee}

AngularFM removes the radial degree of freedom instead of correcting it.
The source in \maineq{4} lies on the joint circles, and each sampling step in \maineq{7} is a rotation.
All generated states therefore have unit norm for any number of sampling steps and any predicted angular velocity (Proposition~\ref{prop:supp_rotation} in \cref{sec:supp_rotation}).
An error $\delta\omega$ of the predicted angular velocity displaces the phase by exactly $\Delta s_k\,\delta\omega$ in a step, with a gain of one that does not depend on the state, and no state has to be decoded from the plane.
This guarantee concerns the geometry of the generated states.
It does not bound the error of the predicted angular velocities themselves.

\section{Additional Derivations}
\label{sec:supp_derivations}

This section complements \mainsec{3}.
The shortest-arc paths, the angular velocity targets, and the rotation steps of AngularFM are an instance of Riemannian flow matching~\cite{chen2024flow} on the product of the joint circles.
We state the properties that are used in the main paper in its notation.

\subsection{Marginal Angular Velocity Field}
\label{sec:supp_marginal}

Consider the source angles $Q_0$ of \maineq{4}, the targets $Q_1$, and the conditional path $\mathbf Z_s=\mathbf u(Q_0+s\boldsymbol\Delta)$ of \maineq{5}.
Along this path, every phase moves with the constant angular velocity $\Delta^{h,i}$.
Since $w(s)>0$ inside the interval, the minimizer of the objective in \maineq{6} is the conditional expectation
\begin{equation}
\boldsymbol\omega^\star(\mathbf Z,s,o)=\mathbb E\left[\boldsymbol\Delta\mid\mathbf Z_s=\mathbf Z,o\right].
\label{eq:supp_omega_star}
\end{equation}
Let $p_s(\cdot\mid o)$ denote the density of the phases $\mathbf Z_s$ on the torus $\mathcal M$ in angular coordinates.
Averaging the transport of the conditional paths over the pairs $(Q_0,Q_1)$ gives the continuity equation
\begin{equation}
\partial_s p_s+\sum_{h,i}\partial_{q^{h,i}}\!\left(p_s\,\omega^{\star\,h,i}\right)=0 ,
\label{eq:supp_continuity}
\end{equation}
as in conditional flow matching~\cite{lipman2022flow,chen2024flow}.
Under the usual regularity conditions, integrating $\dot Q=\boldsymbol\omega^\star(\mathbf u(Q),s,o)$ from the source distribution thus transports it to the distribution of the demonstrated target phases.

Two properties follow from \cref{eq:supp_omega_star}.
First, the expectation is taken over scalar angular velocities, which form a vector space.
The marginal field is again an angular velocity field and has no radial component.
For the sine--cosine baseline, the same argument yields the average of chord vectors in \cref{eq:supp_vstar}, whose radial component does not vanish in general.
Second, the averaging concerns the velocity field and not the samples, so that multimodal action distributions remain representable.

\subsection{Rotation-Based Sampling}
\label{sec:supp_rotation}

\suppprop{rotation steps}\label{prop:supp_rotation}
{\itshape
For any grid $0=s_0<\cdots<s_N=1$ and any predicted angular velocities, all states of the sampler in \maineq{7} have unit norm.
If the angular velocities equal the conditional target, $\boldsymbol\omega_k=\boldsymbol\Delta$ for all $k$, the sampler reaches the target phases exactly, $\mathbf u(\widehat Q_{s_N})=\mathbf u(Q_1)$, for any grid.}

\suppproof
Each step multiplies the phase vector by the rotation $R(\Delta s_k\,\omega_k^{h,i})$, which is orthogonal.
With $\boldsymbol\omega_k=\boldsymbol\Delta$, the steps add up to $\widehat Q_{s_N}=Q_0+\boldsymbol\Delta\sum_k\Delta s_k=Q_0+\boldsymbol\Delta$, and $\mathbf u(Q_0+\boldsymbol\Delta)=\mathbf u(Q_1)$ by the definition of $\boldsymbol\Delta$.
\suppqed

Neither property holds for an Euler step in the plane.
A tangential step $\mathbf z+h\,\omega K\mathbf z$ has the squared norm $1+h^2\omega^2$.
After $N$ uniform steps with a constant angular velocity $\omega$, the radius has grown to $(1+\omega^2/N^2)^{N/2}\approx1+\omega^2/(2N)$, which is 1.13 for $\omega=\pi/2$ and $N=10$.
Projecting back to the circle after each step restores the norm, but the phase then advances by $\arctan(h\omega)$ instead of $h\omega$.
For a velocity with a radial component $\alpha$, the advance is $\operatorname{atan2}(h\omega,1+h\alpha)$ and depends on the radial output as well.

\subsection{State-Anchored Source}
\label{sec:supp_source}

We compare the state-anchored source of \maineq{4} with a uniform source on the circle for one joint and one future step.
Let $d=\operatorname{ang}(\mathbf u(q_{\mathrm{cur}}),\mathbf u(q_1))$ be the displacement from the current phase to the target phase.

\paragraph{Path length.}
For a uniform source, the angular displacement $\Delta$ is uniform on $[-\pi,\pi)$ for every target, so that $\mathbb E[\Delta^2]=\pi^2/3\approx3.29$.
For the anchored source, $\Delta=\operatorname{wrap}(d-\epsilon)$ with $\epsilon\sim\mathcal N(0,\kappa^{-1})$, and
\begin{equation}
\mathbb E\left[\Delta^2\right]\le d^2+\kappa^{-1},
\label{eq:supp_transport}
\end{equation}
since wrapping never increases the magnitude.
The bound is nearly attained unless $|d|$ is close to $\pi$ or $\kappa$ is small.
The anchored paths are shorter than the uniform ones on average whenever $|d|<(\pi^2/3-\kappa^{-1})^{1/2}$, which is 1.67\,rad for $\kappa=2$ and 1.74\,rad for $\kappa=4$.
For $\kappa=4$ and $d=0.5$\,rad, the mean squared displacement is 0.50, compared with 3.29.

\paragraph{Antipodal discontinuity.}
As a function of the source phase, the target $\Delta$ jumps by $2\pi$ where the source phase is antipodal to the target phase.
A source sample lies beyond this point, so that its shortest arc winds against the direct displacement, with probability
\begin{equation}
P=\Phi_c\!\left((\pi-|d|)\sqrt\kappa\right)+\Phi_c\!\left((\pi+|d|)\sqrt\kappa\right),
\label{eq:supp_cut}
\end{equation}
where $\Phi_c$ is the complementary distribution function of the standard normal distribution.
For $d=1$\,rad, this probability is $1.6\times10^{-2}$ for $\kappa=1$, $1.2\times10^{-3}$ for $\kappa=2$, and $9.2\times10^{-6}$ for $\kappa=4$.
A uniform source, in contrast, has the constant density $1/(2\pi)$ at the discontinuity for every target.
In the demonstrated chunks of the rotation-sensitive suite, only 0.41\% of the targets are more than $150^\circ$ away from the current joint angle, and 2.3\% are more than $90^\circ$ away.
The anchored source therefore places almost no mass near the discontinuity, whereas a small precision $\kappa$ moves mass toward it.

\subsection{Sequential Phase Lifting}
\label{sec:supp_lifting}

\suppprop{phase lifting}\label{prop:supp_lifting}
{\itshape
Let $q^0=q_{\mathrm{cur}}$ and let $q^1,\ldots,q^H$ be real-valued joint positions with $|q^h-q^{h-1}|<\pi$ for all $h$.
Given only the phases $\mathbf u(q^h)$, the reconstruction of \maineq{8} returns $\bar q^h=q^h$ for all $h$.}

\suppproof
By induction, assume $\bar q^{h-1}=q^{h-1}$.
The increment $\operatorname{ang}(\mathbf u(q^{h-1}),\mathbf u(q^h))$ is the representative of $q^h-q^{h-1}$ in $[-\pi,\pi)$, which is $q^h-q^{h-1}$ itself by assumption.
Hence $\bar q^h=q^{h-1}+(q^h-q^{h-1})=q^h$.
\suppqed

The reconstruction is therefore exact, including the number of turns, as long as a joint moves by less than half a turn between consecutive targets.
In the rotation-sensitive suite, the largest rotation between consecutive recorded frames is 0.295\,rad, far below this limit.
The same reconstruction is applied to the angles decoded by the sine--cosine baseline.

\begin{table*}[t]
\centering
\small
\caption{Real-world tasks, language instructions, and numbers of teleoperated demonstrations.}
\label{tab:supp_real_tasks}
\setlength{\tabcolsep}{6pt}
\begin{tabular}{ll|l|c}
\toprule
Robot & Task & Language instruction & Demonstrations \\
\midrule
\multirow{3}{*}{Franka}
 & Shoe placement  & ``Place the shoes correctly into the box.'' & 60 \\
 & Block insertion & ``Please insert the yellow block into the hole.'' & 70 \\
 & Block stacking  & ``Stack the blocks from bottom to top in the order of red, yellow, and green.'' & 50 \\
\midrule
\multirow{3}{*}{AgileX}
 & Cup inversion   & ``Place the cup upside down on the drying rack.'' & 50 \\
 & Plate insertion & ``Place the plate in the drying rack.'' & 60 \\
 & Rack flipping   & ``Please turn the drying rack upside down.'' & 50 \\
\bottomrule
\end{tabular}
\end{table*}

\section{Additional Real-World Details and Results}
\label{sec:supp_real}

This section complements \mainsec{4.4}.

\subsection{Robots and Demonstrations}
\label{sec:supp_real_data}

\paragraph{Franka.}
The single-arm platform is a 7-DoF Franka FR3 with a Robotiq 2F-85 gripper.
Demonstrations are collected by teleoperation with a GELLO leader arm~\cite{wu2024gello}.
Two RealSense D435i cameras provide a global view and a wrist view.
Joint states, leader targets, and images are recorded at 30\,Hz.
The action targets are the leader joint positions of the seven arm joints and a binary gripper command, so that $n=7$ and $g=1$.
Each view is center-cropped and resized to $224\times224$, and the two views are concatenated side by side.

\paragraph{AgileX.}
The dual-arm platform is an AgileX Cobot Magic with two 6-DoF arms and parallel grippers, teleoperated through its leader arms.
A head camera and two wrist cameras are recorded at 30\,Hz together with the joint positions.
The follower arms lag the leader arms by about four frames, and we use the measured follower joint positions as action targets.
The joint layout matches the simulated AgileX robot of RoboTwin~2.0, with $n=12$ and $g=2$, and the camera views are arranged as in \cref{sec:supp_arch}.

\paragraph{Tasks.}
\cref{tab:supp_real_tasks} lists the six tasks with their language instructions and numbers of demonstrations.
We train one policy per robot on its three tasks, conditioned on the language instruction.

\subsection{Fine-Tuning}
\label{sec:supp_real_training}

Both policies use $\kappa=2$.
Fine-tuning has two stages.
For both robots, the first stage trains only the three linear layers of the action interface (\cref{sec:supp_arch}) for 1k steps with a learning rate of $3\times10^{-4}$, while the transformer blocks are frozen and the video loss is disabled.
The second stage trains all parameters for 5k steps with a learning rate of $3\times10^{-5}$ and a global batch size of 256.
It applies random image shifts of up to 8 pixels and color jitter with a strength of 0.1, and it maintains an exponential moving average of the weights with a decay of 0.999.

\subsection{Evaluation Protocol}
\label{sec:supp_real_eval}

Every method is evaluated with ten trials per task, and the object layout is varied between trials.
A trial counts as a success if the instructed goal state is reached.
For shoe placement, both shoes lie side by side in the box.
For block insertion, the yellow block is seated upright in the hole.
For block stacking, the three blocks remain stacked in the instructed order after the gripper has released the last block.
For cup inversion, the cup rests upside down on the drying rack.
For plate insertion, the plate stands upright in a slot of the rack.
For rack flipping, the rack is lifted by both arms and turned upside down without being dropped.
\maintab{4} reports the number of successful trials.

\subsection{Qualitative Results}
\label{sec:supp_real_qualitative}

\begin{figure*}[p]
  \centering
  \suppgraphic[width=\linewidth]{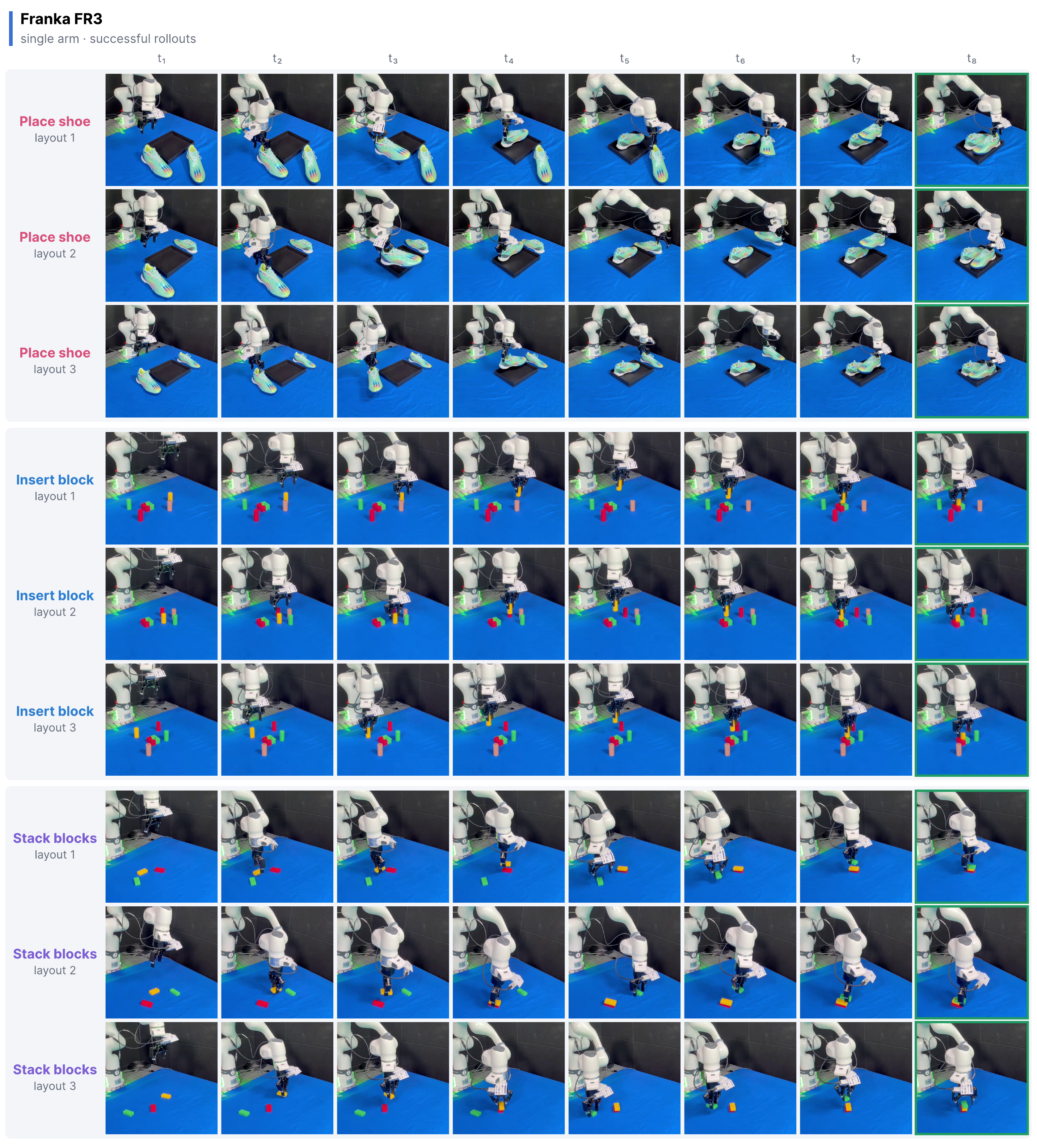}
  \caption{Successful rollouts of AngularFM on the Franka FR3 for shoe placement, block insertion, and block stacking, under three object layouts each. Every row shows eight frames ($t_1$--$t_8$) of one trial, with the final state highlighted in green.}
  \label{fig:supp_franka_success}
\end{figure*}

\begin{figure*}[p]
  \centering
  \suppgraphic[width=\linewidth]{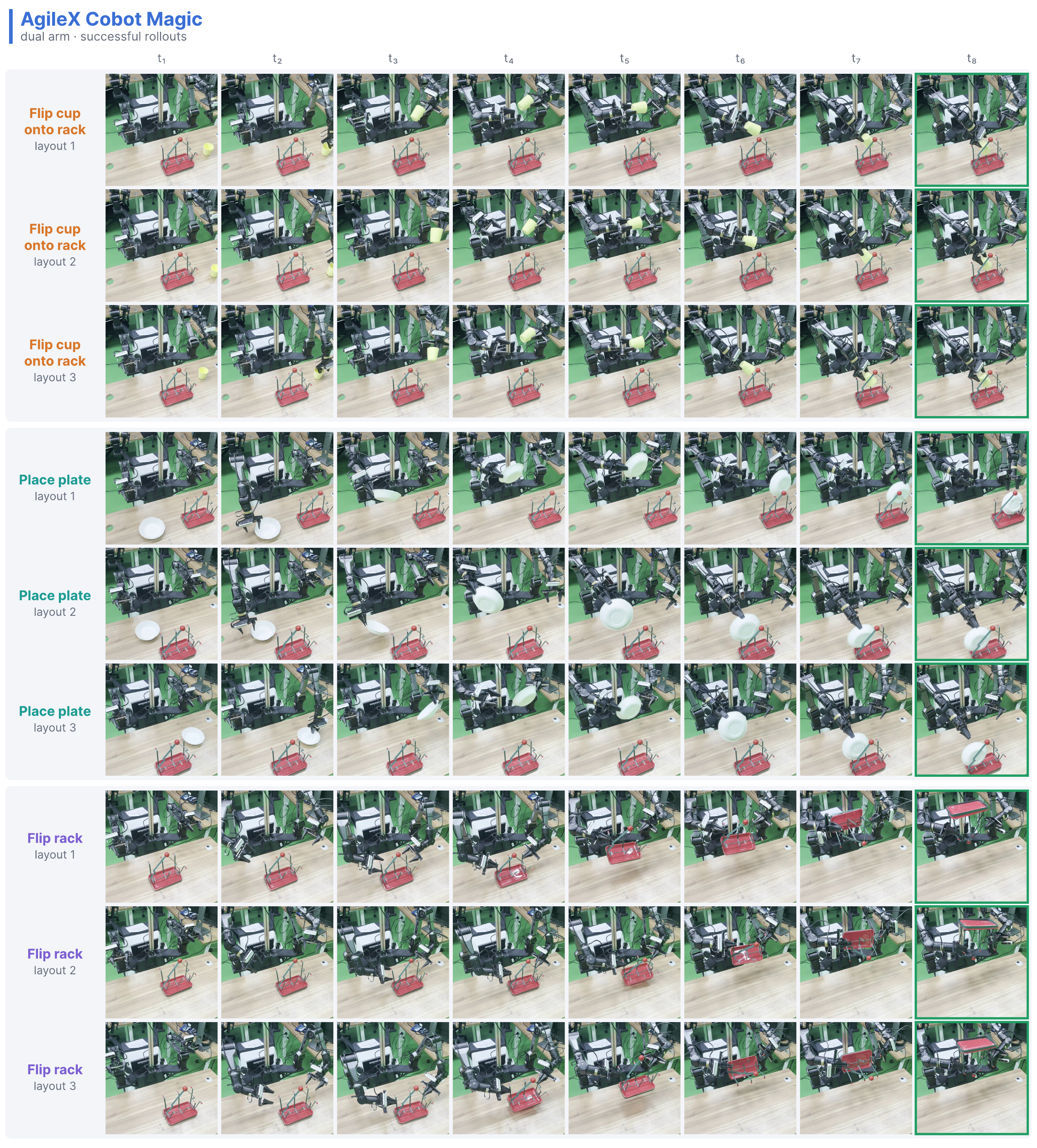}
  \caption{Successful rollouts of AngularFM on the AgileX Cobot Magic for cup inversion with handover, plate insertion, and bimanual rack flipping, under three object layouts each. Every row shows eight frames ($t_1$--$t_8$) of one trial, with the final state highlighted in green.}
  \label{fig:supp_agilex_success}
\end{figure*}

\begin{figure*}[p]
  \centering
  \suppgraphic[width=\linewidth]{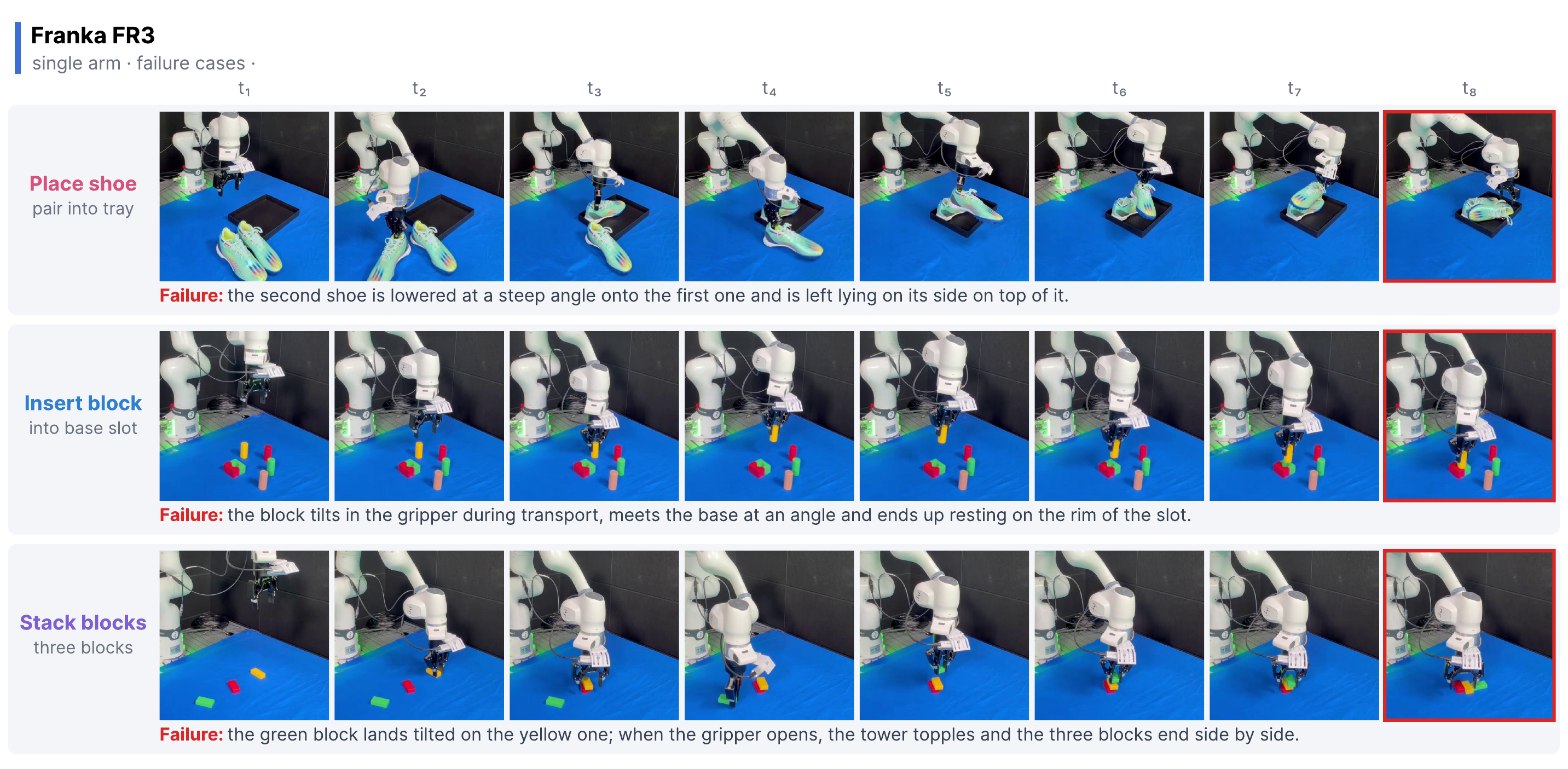}
  \caption{Unsuccessful trials of AngularFM on the Franka FR3, one for each task. Every row shows eight frames ($t_1$--$t_8$), with the final state highlighted in red and the outcome described below the frames.}
  \label{fig:supp_franka_failure}
\end{figure*}

\begin{figure*}[p]
  \centering
  \suppgraphic[width=\linewidth]{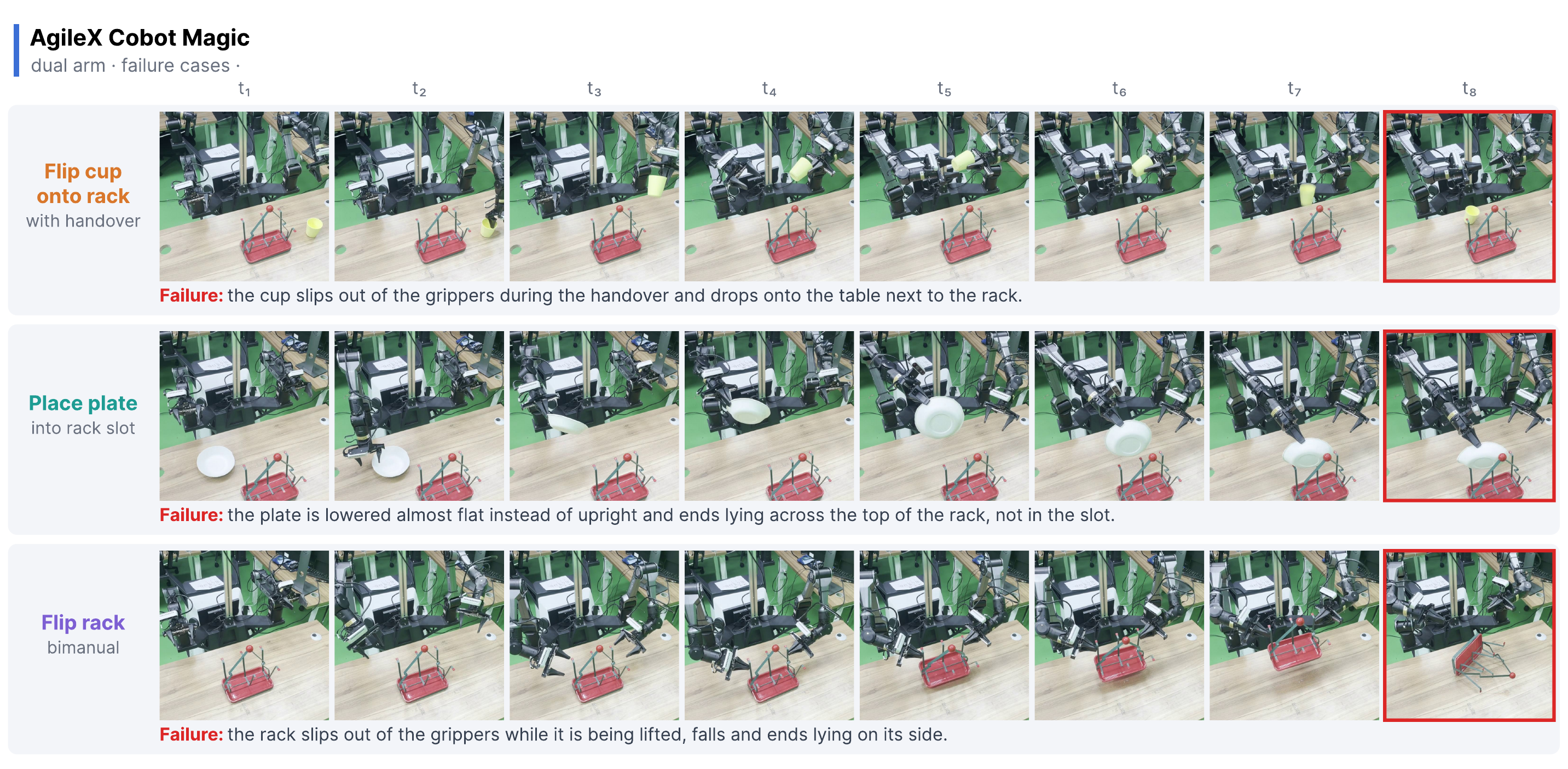}
  \caption{Unsuccessful trials of AngularFM on the AgileX Cobot Magic, one for each task. Every row shows eight frames ($t_1$--$t_8$), with the final state highlighted in red and the outcome described below the frames.}
  \label{fig:supp_agilex_failure}
\end{figure*}

\cref{fig:supp_franka_success,fig:supp_agilex_success} extend \mainfig{3} and show rollouts of AngularFM under three object layouts for each of the six tasks, with eight frames per rollout.
\cref{fig:supp_franka_failure,fig:supp_agilex_failure} show one unsuccessful trial for each task.

\paragraph{Successful rollouts.}
On the Franka (\cref{fig:supp_franka_success}), the three layouts of a task differ in the positions and orientations of the objects.
In shoe placement, the two shoes start at different positions and headings around the box.
The policy grasps one shoe, turns it to the orientation of the box, places it, and repeats the sequence for the second shoe, so that both shoes end side by side.
The required wrist rotation therefore differs between the layouts and between the two shoes of one layout.
In block insertion, the yellow cylindrical block and the two notched blocks that form the hole stand at different positions among distractor blocks.
The arm grasps the yellow block, carries it above the hole, and seats it upright.
In block stacking, the policy places the yellow block on the red one and then the green block on top, and each placement has to be aligned with the block below.

On the AgileX (\cref{fig:supp_agilex_success}), all three tasks reorient an object relative to the drying rack.
In cup inversion, the first arm picks up the cup and turns it sideways, and the second arm takes it over and places it upside down on the rack.
In plate insertion, the plate is picked up from the table at its rim, turned from a flat to an upright pose, and lowered into a slot of the rack.
In rack flipping, both arms grasp the rack at its two ends, lift it, and rotate it jointly by about half a turn.
Across the layouts, the same policy completes the grasping, reorientation, and placement stages for different object poses.

\paragraph{Failure cases.}
AngularFM does not complete 13 of the 60 trials in \maintab{4}, and \cref{fig:supp_franka_failure,fig:supp_agilex_failure} show one such trial for each task.
In all six trials, the policy reaches and grasps the object and moves it toward the target, and the trial is lost in a later, contact-rich phase.
In three trials, the object moves inside the gripper: the yellow block tilts during transport and is left on the rim of the hole, the cup slips during the handover, and the rack slips while it is lifted.
In the other three trials, the object is released in an imprecise pose: the second shoe is lowered at a steep angle and comes to rest on the first shoe, the plate is lowered before it is fully upright and ends on top of the rack, and the last block lands tilted, so that the stack collapses when the gripper opens.

These outcomes arise in the phases with the smallest tolerances, and they are not specific to AngularFM.
Failures of the same kinds were also observed for the compared methods, for each of which \maintab{4} lists unsuccessful trials on every task.
A slip or a tilt of the object inside the gripper is difficult to perceive from the camera views and can only be corrected at the next re-planning step, and a release with a small pose error cannot be undone without a recovery behavior such as re-grasping.
We therefore regard these cases as remaining challenges of real-world manipulation that current visuomotor policies share.
Tactile feedback and learned recovery behaviors could address them and are complementary to the joint-angle formulation studied in this paper.

\end{document}